%% file: 0_main.tex
\documentclass[manuscript]{acmart}

\usepackage{xr}
\usepackage{amsmath}

\usepackage{amssymb}
\usepackage{pifont}

\usepackage{subfig}
\usepackage{comment}
\usepackage{soul}
\usepackage{algorithmic}
\usepackage{hyperref}
\usepackage{booktabs}
\usepackage{threeparttable}
\usepackage{mathrsfs}
\usepackage{amsfonts}
\usepackage[ruled]{algorithm2e}

\usepackage{multirow}
\usepackage{graphicx}
\usepackage{color}
\usepackage{balance}
\usepackage{caption}
\usepackage{adjustbox}
\usepackage{soul}
\setstcolor{red}
\usepackage{array}
\usepackage{utfsym}
\usepackage{multirow}
\usepackage{enumitem}

\newcommand{\boldcheckmark}{\textcolor{black}{\pmb{\checkmark}}}  
\newcommand{\gcmark}{\textcolor{green}{\checkmark}}  
\newcommand{\rxmark}{\textcolor{red}{\ding{55}}}     

\usepackage{color}
\usepackage{tikz}
\usepackage[edges]{forest}
\definecolor{hidden-draw}{RGB}{0,0,0}  
\definecolor{hidden-pink}{rgb}{0.98, 0.94, 0.75}  
\definecolor{level0}{rgb}{0.67, 0.88, 0.69}
\definecolor{level1}{rgb}{0.98, 0.92, 0.84}
\definecolor{level2}{rgb}{0.8, 0.8, 1.0}
\definecolor{level3}{rgb}{1.0, 0.71, 0.76}

\newcommand{\definitionautorefname}{Definition}

\newcommand*{\shortautoref}[1]{%
  \begingroup
    \def\sectionautorefname{Sec.}%
    \def\subsectionautorefname{Sec.}%
    \def\figureautorefname{Fig.}%
    \def\tableautorefname{Tab.}%
    \def\equationautorefname{Eq.}%
    \def\subfigureautorefname{Fig.}%
    \def\definitionautorefname{Def.}%
    \autoref{#1}%
  \endgroup
}

\usepackage{tikz}

\newtheorem{definition}{Definition}
\usepackage{makecell}

\begin{document}

\title{Large Models for Time Series and Spatio-Temporal Data: A Survey and Outlook}

\author{Ming Jin} 
\email{mingjinedu@gmail.com} 
\affiliation{%
  \institution{Griffith University} 
  \city{Brisbane} 
  \state{Queensland} 
  \country{Australia} 
}

\author{Yaxuan Kong}
\email{yaxuan.kong@eng.ox.ac.uk}
\affiliation{%
  \institution{University of Oxford}
  \city{Oxford}
  \country{United Kingdom}
}

\author{Yuxuan Liang}
\email{yuxliang@outlook.com}
\affiliation{%
  \institution{Hong Kong University of Science and Technology (Guangzhou)}
  \city{Guangzhou}
  \country{China}
}

\author{Chaoli Zhang} 
\email{chaolizcl@zjnu.edu.cn} 
\affiliation{%
  \institution{Zhejiang Normal University} 
  \city{Jinhua} 
  \state{Zhejiang} 
  \country{China} }

\author{Siqiao Xue}
\email{siqiao.xsq@antgroup.com}
\affiliation{%
  \institution{Ant Group}
  \city{Hangzhou}
  \state{Zhejiang}
  \country{China}
}

\author{Xue Wang}
\email{xue.w@alibaba-inc.com}
\affiliation{%
  \institution{Alibaba Group}
  \city{Hangzhou}
  \state{Zhejiang}
  \country{China}
}

\author{James Zhang}
\email{jameszhang@deloitte.com}
\affiliation{%
  \institution{Deloitte Service LLP}
  \city{New York}
  \country{USA}
}

\author{Yi Wang}
\email{yi-wang@eee.hku.hk}
\affiliation{%
  \institution{The University of Hong Kong}
  \city{Hong Kong}
  \country{China}
}

\author{Haifeng Chen}
\email{haifeng@nec-labs.com}
\affiliation{%
  \institution{NEC Laboratories America}
  \city{Princeton}
  \state{New Jersey}
  \country{USA}
}

\author{Xiaoli Li}
\email{xlli@i2r.a-star.edu.sg}
\affiliation{%
  \institution{A*STAR}
  \city{Singapore}
  \country{Singapore}
}

\author{Vincent S. Tseng}
\email{vtseng@cs.nycu.edu.tw}
\affiliation{%
  \institution{National Yang Ming Chiao Tung University}
  \city{Hsinchu}
  \country{Taiwan}
}

\author{Yu Zheng}
\email{msyuzheng@outlook.com}
\affiliation{%
  \institution{JD Technology}
  \city{Beijing}
  \country{China}
}

\author{Lei Chen}
\email{leichen@cse.ust.hk}
\affiliation{%
  \institution{Hong Kong University of Science and Technology (Guangzhou)}
  \city{Guangzhou}
  \country{China}
}

\author{Hui Xiong}
\email{xionghui@ust.hk}
\affiliation{%
  \institution{Hong Kong University of Science and Technology (Guangzhou)}
  \city{Guangzhou}
  \country{China}
}

\author{Shirui Pan}
\authornotemark[1]
\email{s.pan@griffith.edu.au}
\affiliation{%
  \institution{Griffith University}
  \city{Gold Coast}
  \state{Queensland}
  \country{Australia}
}

\author{Qingsong Wen}
\authornote{Correspondence to: Shirui Pan <s.pan@griffith.edu.au> and Qingsong Wen <qingsongedu@gmail.com>.}
\email{qingsongedu@gmail.com}
\affiliation{%
  \institution{Squirrel Ai Learning}
  \city{Bellevue}
  \state{Washington}
  \country{USA}
}

\renewcommand{\shortauthors}{Jin et al.}

\begin{abstract}
  \input{plain_abstract.txt}
\end{abstract}






\maketitle

\input{sections/1_introduction}
\input{sections/2_preliminaries}
\input{sections/3_taxonomy}

\input{sections/4_lm4ts}
\input{sections/5_lm4st}
\input{sections/6_resource_and_application}
\input{sections/7_future_direction}
\input{sections/8_conclusion}



\bibliographystyle{ACM-Reference-Format}
\bibliography{reference}

\appendix

\end{document}

%% file: plain_abstract.txt
Temporal data — including time series and spatio-temporal data — are pervasive in real-world applications. Generated in massive volumes by physical and virtual sensors, they record dynamic system behaviors and enable a wide range of downstream tasks. Effectively analyzing such data is crucial to unlocking their rich information content. Recent advances in large language models and other foundation models have accelerated their use in time series and spatio-temporal data mining. These approaches not only improve pattern recognition and reasoning across diverse domains but also support progress toward artificial general intelligence that can understand and process temporal data. In this survey, we present a comprehensive, up-to-date review of large models tailored or adapted for time series and spatio-temporal data along four dimensions: data types, model categories, model scopes, and application areas/tasks. We organize existing work into two main groups: large models for time series analysis (LM4TS) and for spatio-temporal data mining (LM4STD), and further distinguish general-purpose from domain-specific models. We also curate related resources, including datasets, model implementations, and tools, organized by major application areas. Overall, this survey consolidates recent advances and highlights foundations, applications, resources, and open research opportunities in large model–centric temporal data analysis.

%% file: sections/1_introduction.tex
\section{Introduction} \label{sec:introduction}
Large models, specifically referred to as \textit{large language models} (LLMs) and \textit{pre-trained foundation models} (PFMs), have witnessed remarkable success across a multitude of tasks and domains, such as natural language processing (NLP)~\cite{zhao2023survey}, computer vision (CV)~\cite{awais2023foundational}, and various other interdisciplinary areas~\cite{latif2023sparks, singhal2023large, mirchandani2023large}. Initially, LLMs are developed to serve as pre-trained language foundation models for solving different natural language tasks, such as text classification~\cite{kant2018practical}, question answering~\cite{su2019generalizing}, and machine translation~\cite{zhang2023prompting}. However, the emergent abilities of LLMs~\cite{wei2022emergent} to learn intricate semantic and knowledge representations from large-scale text corpora to reason on various tasks have profoundly inspired the community. A case in point is GPT-3~\cite{brown2020language}, endowed with 175 billion parameters, which has exhibited robust few-shot and zero-shot learning capabilities, a feat not achieved by its predecessor, GPT-2~\cite{radford2019language}, which has 1.5 billion parameters. Another example is PaLM~\cite{chowdhery2022palm,anil2023palm} with even more parameters, showcasing remarkable performance on language understanding, general reasoning, and even code-related tasks. The meteoric rise of LLMs has concurrently catalyzed and reconfigured the landscape of PFMs, though the underpinnings such as deep neural networks, self-supervised learning~\cite{liu2021self}, and transfer learning~\cite{weiss2016survey} have been well studied for years. A salient illustration is the emergence of vision-language models (VLMs)~\cite{awais2023foundational} to reason on both visual and textual data, showing promising results in diverse tasks like image captioning~\cite{hossain2019comprehensive}, visual question answering~\cite{kafle2017visual}, and commonsense reasoning~\cite{zellers2019recognition}. Recently, the influence of large models has extended into other sectors, such as audio and speech analysis~\cite{latif2023sparks}, covering an expansive array of modalities (e.g., audio, images, and text) and tasks. Given the remarkable achievements of large models in these diverse fields, an intriguing question emerges: \textbf{\textit{can large models be effectively employed to analyze time series and spatio-temporal data?}}

\begin{figure*}[t]
    \centering
    \includegraphics[width=0.8\textwidth]{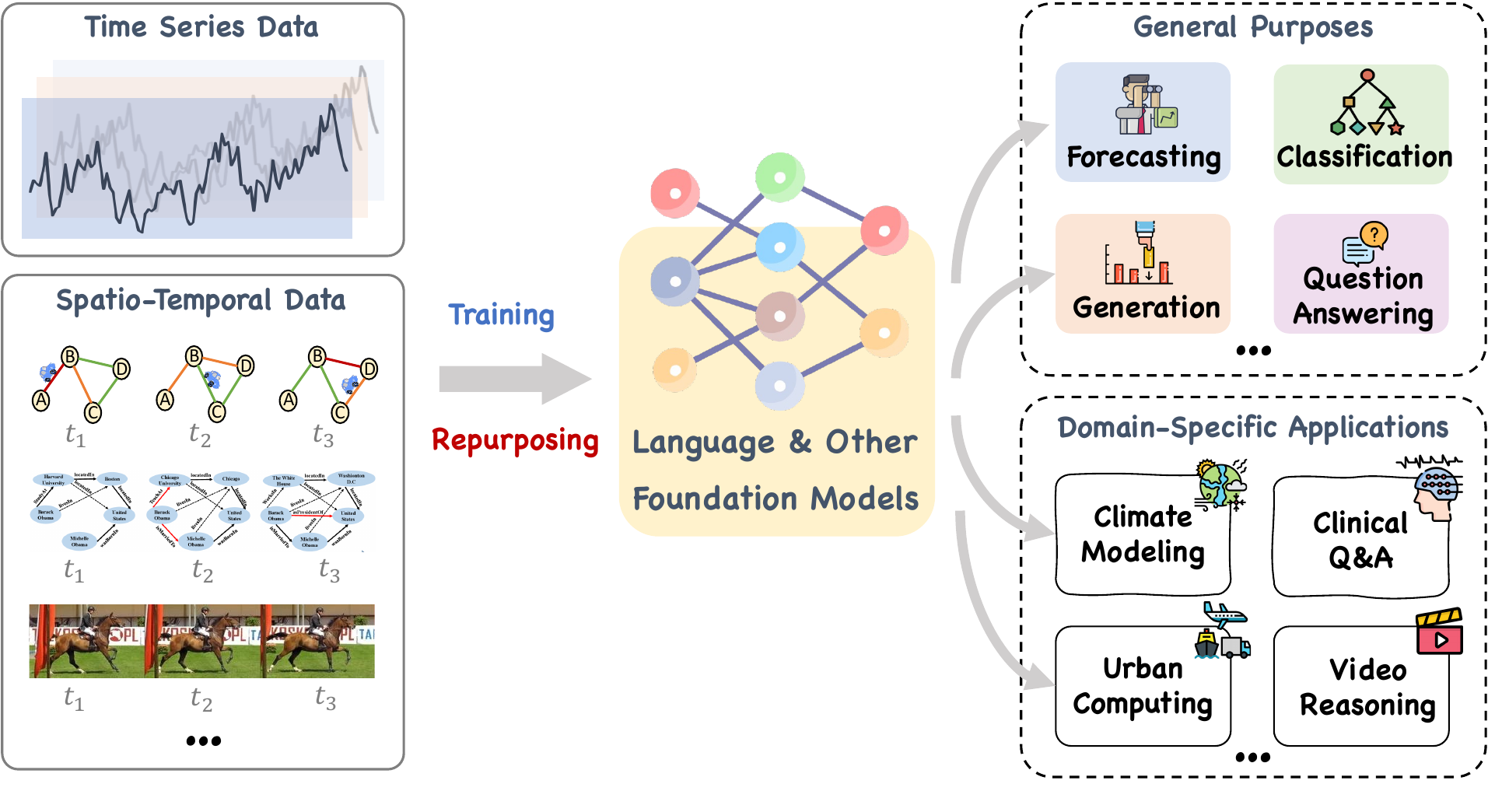}
    \caption{Large models (i.e., language and other related foundation models) can be either trained or adeptly repurposed to handle time series and spatio-temporal data for a range of general-purpose tasks and specialized domain applications.}
    \label{fig:introduction}
\end{figure*}
    
Temporal data, primarily consisting of time series and spatio-temporal data, have long been studied and proven to be indispensable in a myriad of real-world applications, encompassing fields such as geoscience~\cite{karpatne2018machine}, transportation~\cite{jin2023spatio}, energy~\cite{zhu2023eforecaster}, healthcare~\cite{harutyunyan2019multitask}, environment~\cite{xia2023deciphering}, and finance~\cite{tsay2005analysis}. 
Both time series and spatio-temporal data are fundamentally temporal in nature and encapsulate various data modalities (see \shortautoref{sec:time series and spatial-temporal data}). While they share commonalities in modeling temporal dynamics, spatio-temporal data also require the integration of spatial dependencies to precisely capture the interactions within the data. Combining the study of these two broad and intrinsically linked data categories is crucial, as they are both pervasive and can be sourced from a diverse range of platforms, such as sensors~\cite{lazaridis2003capturing,yao2017deepsense}, financial market transactions~\cite{sezer2020financial}, cloud monitoring~\cite{zhou2023ahpa}, and so on. By surveying them together, we can gain a holistic understanding of the inherent dynamics they encapsulate across various systems.

While large models have rapidly reshaped many fields (see \shortautoref{fig:roadmap}), time series and spatio-temporal analysis have historically developed through statistical modeling and compact deep architectures. Classical statistical models date back to early time series studies \cite{yule1926we}. Deep learning later introduced stronger data-driven models based on recurrent neural networks (RNNs)~\cite{hewamalage2021recurrent}, convolutional neural networks (CNNs)~\cite{lea2016temporal}, graph neural networks (GNNs)~\cite{jin2023survey, jin2023spatio}, and Transformers~\cite{selva2023video, wen2022transformers}. However, most of these models are relatively small and task-specific, which limits their ability to support broad transfer, semantic representation, and multi-task reasoning. Earlier self-supervised pre-training approaches, such as TF-C~\cite{zhang2022self}, SimMTM~\cite{dong2023simmtm}, STEP~\cite{shao2022pre}, and AdaMAE~\cite{bandara2023adamae}, improved transferability but mainly served as precursors to the current large-model paradigm.

Despite the remarkable progress in developing large models for time series and spatio-temporal data, the limited availability of high-quality, diverse large-scale datasets still poses a substantial challenge in many domains~\cite{zhou2023one, jin2023timellm}. Even so, recent years have witnessed a surge of successful applications across various tasks and sectors (see \shortautoref{tab:summary}), underscoring the growing potential of foundation models for temporal data analysis. In data-rich scientific domains, Pangu-Weather~\cite{bi2022pangu} and ClimaX~\cite{nguyen2023climax} have revolutionized global climate modeling by delivering unprecedented forecasting accuracy while dramatically reducing computational costs. In the field of urban computing, TFM~\cite{wang2023building} introduced one of the first transportation foundation models. For spatio-temporal video understanding, models such as Valley~\cite{luo2023valley}, LAVILA~\cite{zhao2023learning}, and mPLUG-2~\cite{xu2023mplug} demonstrate how large-scale multimodal pre-training can capture complex temporal dynamics. The trend has also expanded to the time series foundation model (TSFM) space. Representative examples include Chronos~\cite{ansari2024chronos}, Time-MoE~\cite{shi2024time}, and MOIRAI~\cite{woo2024unified}, which aim to build general-purpose models for forecasting and representation learning across diverse domains. Complementary to TSFMs, recent efforts have sought to harness the reasoning and generalization capabilities of LLMs for time series analysis. Early explorations such as PromptCast~\cite{xue2022promptcast} and OFA~\cite{zhou2023one} leveraged LLMs for generic time series analysis through prompting or lightweight fine-tuning, while Time-LLM~\cite{jin2023timellm} proposed reprogramming time series inputs with natural-language prompts to exploit off-the-shelf LLMs. In specialized applications, OpenTSLM~\cite{langer2025opentslm} and MedTsLLM~\cite{chan2024leveraging} exemplify large clinical language models that show promise for medical event prediction. Despite these significant strides and their inherent promise, the integration of large models into time series and spatio-temporal data analysis continues to confront unique challenges that necessitate focused investigation (see \shortautoref{sec:future directions}). This underscores the necessity for a comprehensive review tailored to examine the progress in this burgeoning field.

In this paper, we fulfill this need by providing a unified, comprehensive, and up-to-date review of large models dedicated to time series and spatio-temporal data analysis, encompassing both LLMs and PFMs across different data categories, model scopes, application domains, and representative tasks. Time series and spatio-temporal data are both temporal in nature and often share similarities in analytical methods. Combining them in a single survey allows practitioners to explore the synergies and commonalities between these two areas. Integrating insights from both domains can also foster cross-pollination of ideas. This integration may help researchers transfer strengths from one domain to address challenges in the other and build a more interconnected understanding of large models for temporal data. Our contributions are summarized as follows:

\vspace{-1mm}

\begin{itemize}
    \item \textbf{A Comprehensive and Foundational Survey.} This work offers an extensive and in-depth survey of advancements in large models for time series and spatio-temporal data analysis. By systematically charting the field's landscape and examining the nuances of key methodologies, this paper serves as a foundational resource for researchers and practitioners, delivering a thorough and organized understanding of this dynamic area of study.
    \item \textbf{Unified and Structured Taxonomy.} We introduce a unified and structured taxonomy that categorizes existing research into two principal clusters: large model for time series (LM4TS) and for spatio-temporal data (LM4STD), organized based on data categories. We further divide each cluster into two subgroups, i.e., LLMs and PFMs, according to model types. Subsequent categorizations are conducted through the lenses of model scopes, application domains, and specific tasks. This multi-faceted classification provides the reader with a coherent roadmap for understanding the field from multiple perspectives.
    \item \textbf{Abundant Resource Compilation.} We compile and summarize a wealth of resources in this field, encompassing datasets, open-source implementations, and evaluation benchmarks. Additionally, we outline the practical applications of pertinent large models across various domains. This compilation serves as a valuable reference point for future research and development endeavors.
    \item \textbf{Future Research Opportunities.} We identify and elaborate on multiple promising avenues for future research, encompassing a range of perspectives such as data sources, model architectures, training and inference paradigms, as well as other potential opportunities. This discussion equips the reader with a nuanced understanding of the current state of the field, while also highlighting prospective directions for future investigation.
\end{itemize}

\input{tables/compare-to-other-surveys}

\noindent\textbf{Related Surveys and Differences.} 
Several recent surveys have reviewed temporal data analysis from different perspectives, including foundation models for time series~\cite{liang2024foundation,kottapalli2025foundation,liu2025empowering}, LLM-based time-series methods~\cite{shi2025large}, GNNs and structural inductive biases for spatio-temporal prediction~\cite{capone2025spatio,liang2025foundation,wang2025graph}, video foundation and diffusion models~\cite{madan2024foundation,wang2025survey}, and temporal knowledge graphs~\cite{cai2024survey,zhang2024survey}. However, these reviews typically focus on a single modality or model family and rarely provide an integrated view across time series, spatio-temporal graphs, temporal knowledge graphs, and videos. Motivated by the rapid development of large models in vision~\cite{awais2023foundational}, audio~\cite{latif2023sparks}, and geoscience~\cite{mai2023opportunities}, this survey provides a unified review of large models for time series and spatio-temporal data analysis. As summarized in \shortautoref{table:compare_survey}, our distinctive contribution is an integrated temporal-data perspective: we organize these data types under a unified LM4TS/LM4STD taxonomy and compare LLM repurposing and PFM pre-training across model scope, task type, application domain, and resource availability. This complements specialized surveys by highlighting cross-modality links, methodological differences, and transferable adaptation strategies.
\\

\textbf{Survey Methodology and Scope.}
To enhance the verifiability of the survey coverage, we followed an iterative literature collection process. We searched major scholarly databases and indexing platforms, including ACM Digital Library, IEEE Xplore, SpringerLink, ScienceDirect, Web of Science, Google Scholar, and arXiv. We included representative studies on large models for time series and spatio-temporal data published or publicly released up to the time of submission. The search focused on works related to LLM adaptation, pre-trained foundation models, and large-model-based temporal data analysis across time series, spatio-temporal graphs, temporal knowledge graphs, and videos. We further expanded the candidate set through backward and forward snowballing from related surveys. The final selection prioritized early studies that initiated important research directions, influential works with clear methodological contributions, and peer-reviewed papers in relevant conferences and journals. For rapidly emerging topics, we also considered widely cited papers. We excluded works with only incidental large-model relevance, conventional small-scale neural methods without pre-training or large-model adaptation, and duplicate versions lacking distinct technical contributions.
\\

\noindent\textbf{Paper Organization.} The remainder of this paper is structured as follows: \shortautoref{sec:background} furnishes the reader with essential background knowledge on large models, time series, and spatio-temporal data, as well as associated tasks. \shortautoref{sec:overview and categorization} introduces our unified taxonomy for large models in the context of time series and spatio-temporal data analysis, offering a broad overview of the field before delving into the intricacies of specific approaches, which are elaborated upon in \shortautoref{sec: Large Models for Time Series Data} and \shortautoref{sec: Large Models for Spatio-Temporal Data}. \shortautoref{sec:resources} encapsulates the extensive resources and applications we have compiled pertaining to large models for time series and spatio-temporal data analysis. \shortautoref{sec:future directions} outlines promising avenues for future research in this domain. Finally, we conclude our paper in 
\shortautoref{sec:conclusion}.

%% file: tables/compare-to-other-surveys.tex
\begin{table*}[htbp]
    \centering  
    \small
    \renewcommand\arraystretch{1.08}
    \caption{
        Comparison between this and other related surveys, focusing on domains (i.e., specific vs. general), relevant modalities (e.g., time series, spatio-temporal graphs (STGs), temporal knowledge graphs (TKGs), and videos), and primary areas of focus (i.e., small-scale pre-training and fine-tuning (P\&F), large language models (LLMs), and pre-trained foundation models (PFMs)).
        }
    \begin{threeparttable}
    \begin{tabular}{p{0.16\textwidth} >{\centering\arraybackslash}p{0.03\textwidth} | >{\centering\arraybackslash}p{0.06\textwidth} >{\centering\arraybackslash}p{0.07\textwidth} | >{\centering\arraybackslash}p{0.1\textwidth} >{\centering\arraybackslash}p{0.05\textwidth} >{\centering\arraybackslash}p{0.05\textwidth} >{\centering\arraybackslash}p{0.05\textwidth} | >{\centering\arraybackslash}p{0.05\textwidth} >{\centering\arraybackslash}p{0.05\textwidth} >{\centering\arraybackslash}p{0.05\textwidth}}
    \toprule\toprule
    \multicolumn{1}{c}{\multirow{2}{*}{\textbf{Survey}}} & \multicolumn{1}{c|}{\multirow{2}{*}{\textbf{Year}}} & \multicolumn{2}{c|}{\textbf{Domain}} & \multicolumn{4}{c|}{\textbf{Modality}} & \multicolumn{3}{c}{\textbf{Focus}}   \\ 
    \cline{3-4} \cline{5-8} \cline{9-11} 
    \multicolumn{2}{c|}{} & \textbf{Specific}        & \textbf{General}        & \textbf{Time Series} & \textbf{STGs} & \textbf{TKGs} & \textbf{Videos} & \textbf{P\&F} & \textbf{LLMs} & \textbf{PFMs} \\ \hline
    
    Ma \textit{et al.}~\cite{ma2023survey} & 2023 &  & \boldcheckmark & \gcmark & \rxmark & \rxmark & \rxmark & \gcmark & \rxmark & \rxmark \\

    Latif \textit{et al.}~\cite{latif2023sparks} & 2023 & \boldcheckmark &  & \rxmark & \rxmark & \rxmark & \rxmark & \rxmark & \gcmark & \gcmark \\

    Cai \textit{et al.}~\cite{cai2024survey} & 2024 &  & \boldcheckmark & \rxmark & \rxmark & \gcmark & \rxmark & \gcmark & \gcmark & \rxmark \\

    Zhang \textit{et al.}~\cite{zhang2024survey} & 2024 &  & \boldcheckmark & \rxmark & \rxmark & \gcmark & \rxmark & \gcmark & \gcmark & \rxmark \\

    Madan \textit{et al.}~\cite{madan2024foundation} & 2024 & \boldcheckmark &  & \rxmark & \rxmark & \rxmark & \gcmark & \gcmark & \gcmark & \gcmark \\

    Capone \textit{et al.}~\cite{capone2025spatio} & 2025 &  & \boldcheckmark & \rxmark & \gcmark & \rxmark & \rxmark & \rxmark & \rxmark & \rxmark \\
    
    Wang \textit{et al.}~\cite{wang2025graph} & 2025 &  & \boldcheckmark & \rxmark & \gcmark & \rxmark & \rxmark & \gcmark & \gcmark & \gcmark \\

    Kottapalli \textit{et al.}~\cite{kottapalli2025foundation} & 2025 &  & \boldcheckmark & \gcmark & \rxmark & \rxmark & \rxmark & \gcmark & \gcmark & \gcmark \\

    Shi \textit{et al.}~\cite{shi2025large} & 2025 &  & \boldcheckmark & \gcmark & \gcmark & \rxmark & \rxmark & \gcmark & \gcmark & \rxmark \\

    Liang \textit{et al.}~\cite{liang2025foundation} & 2025 &  & \boldcheckmark & \gcmark & \gcmark & \gcmark & \rxmark & \gcmark & \gcmark & \gcmark \\ \hline

    \textbf{This Survey} & 2025 &  & \boldcheckmark & \gcmark & \gcmark & \gcmark & \gcmark & \gcmark & \gcmark & \gcmark \\ \bottomrule\bottomrule

    \end{tabular}
    \end{threeparttable}
\label{table:compare_survey}
\end{table*}

%% file: sections/2_preliminaries.tex
\section{Background} \label{sec:background}

\begin{figure*}[t]
    \centering
    \includegraphics[width=0.9\textwidth]{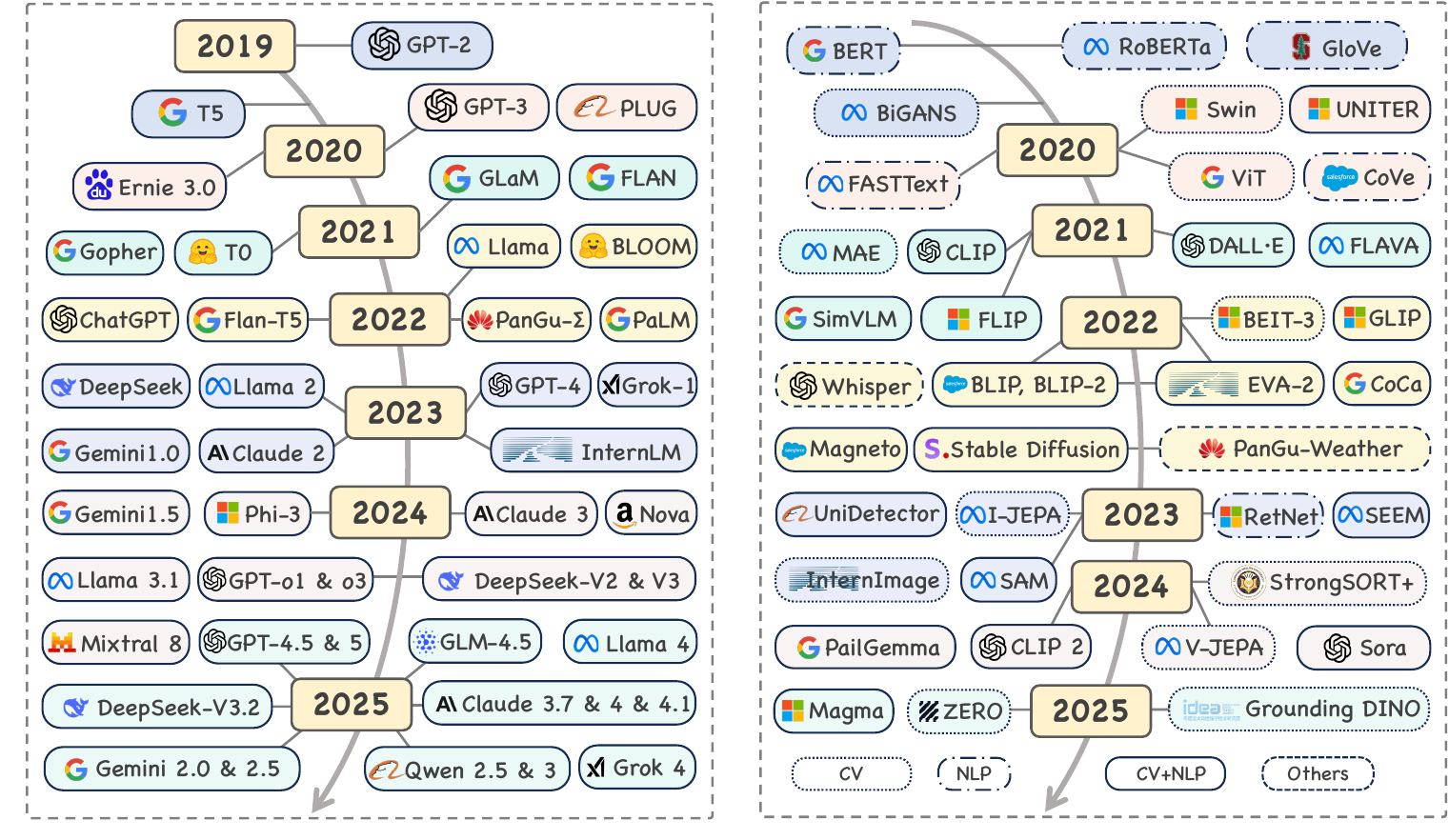}
    \vspace{-5pt}
    \caption{Roadmaps of representative large language models (\textbf{left}) and other foundation models (\textbf{right}).}
    \vspace{-10pt}
    \label{fig:roadmap}
\end{figure*}

This work reviews recent advances in adopting large models for solving time series and spatio-temporal data tasks. Specifically, we focus on two operational groups of large models commonly used in temporal data analysis: LLMs and PFMs. In the broad sense, LLMs can be regarded as a subclass of PFMs; however, in this survey, we distinguish them according to their role in existing studies. LLMs mainly refer to language-centered foundation models repurposed for time series or spatio-temporal tasks through prompting, reprogramming, or adaptation, whereas PFMs refer to pre-trained or foundation models designed or adapted as general-purpose backbones for temporal, spatial, relational, visual, or multimodal data. Accordingly, methods categorized as LLM4TS or LLM4STD are included when a large model serves as a central backbone, interface, or reasoning module, but they are not necessarily claimed to be newly trained standalone foundation models. In this section, we introduce these two groups and discuss their differences and connections. To improve clarity, we also provide a brief roadmap of large models in \shortautoref{fig:roadmap}, which summarizes their development and sheds light on main research interests. Then, we define time series and spatio-temporal data, along with their representative tasks across different domains.

\subsection{Large Language Models} \label{sec:large language models}
Language modeling forms the basis of many NLP tasks, and LLMs are originally developed to enhance language modeling performance. Compared with conventional language models, such as \textit{neural language models} (NLMs)~\cite{mikolov2010recurrent} and small-sized \textit{pre-trained language models} (PLMs)~\cite{devlin2018bert, lewis2019bart, radford2019language}, LLMs are known for their emergent ability~\cite{wei2022emergent} in solving various complex tasks~\cite{zhao2023survey} through in-context learning~\cite{dong2022survey}, reshaping the way we use AI. Recently, with the development of \textit{multimodal large language models} (MLLMs)~\cite{yin2023survey}, the downstream tasks of LLMs have extended far beyond traditional natural language scope~\cite{menon2022visual, lopez2023can} and cannot be easily solved by smaller PLMs. Upon the recent progress of modeling time series and spatio-temporal data with LLMs, we categorize them into two main types: \textit{embedding-visible LLMs} and \textit{embedding-invisible LLMs}, similar to \cite{chen2023exploring, zhao2023survey}. The former type of LLMs is usually open-sourced with publicly accessible inner states. Typical examples include BLOOM~\cite{muennighoff2022crosslingual}, Llama~\cite{touvron2023llama,touvron2023llama2}, Alpaca~\cite{taori2023alpaca}, Vicuna~\cite{chiang2023vicuna}, and Falcon~\cite{almazrouei2023falcon}. This branch of LLMs typically allows for the fine-tuning on different target tasks and shows promising few-shot and zero-shot capabilities without additional retraining.
The latter type of LLMs is usually closed-sourced with no publicly available inner states, such as PaLM~\cite{chowdhery2022palm}, ChatGPT\footnote{https://openai.com/blog/chatgpt/}, and GPT-4~\cite{OpenAI2023GPT4TR}, which are normally inferenced via prompting in API calls. A brief roadmap of LLMs is depicted in \shortautoref{fig:roadmap} (\textbf{left})~\cite{zhao2023survey}.

Witnessing the big success of MLLMs, one of our primary interests is investigating how to adapt LLMs for solving time series and spatio-temporal data analytical tasks, which can generally be achieved by either 
\textit{multimodal re-purposing} or \textit{API-based prompting} in the existing literature.
The former approach is normally used for activating the task-related capabilities of embedding-visible LLMs by aligning different modalities in the target and pre-training (source) tasks. This is closely related to the fine-tuning of LLMs (e.g., adapter tuning~\cite{houlsby2019parameter} and low-rank adaptation~\cite{hu2021lora}) and model reprogramming~\cite{chen2022model}, depending on whether LLMs are fine-tuned or frozen during the adaptation process. Two typical examples are OFA~\cite{zhou2023one} and Time-LLM~\cite{jin2023timellm} detailed in \shortautoref{sec: LLMs in Time Series}.
The latter approach, on the other hand, is more straightforward by wrapping the target modalities into natural language prompts and feeding them into LLMs for generative inference. This is similar to the black-box tuning for language-model-as-a-service (LMaaS)~\cite{sun2022black}. Examples include LLMTime~\cite{gruver2023large} (\shortautoref{sec: LLMs in Time Series}), PPT~\cite{xu2023pre} (\shortautoref{sec: temporal knowledge graphs}), and VideoChat~\cite{li2023videochat} (\shortautoref{sec: video data}).
Evidence shows that both paradigms are applicable and have shown promising results in different target tasks across various domains~\cite{chen2023exploring, tsimpoukelli2021multimodal}, including learning on time series~\cite{gruver2023large,zhou2023one,jin2023timellm} and spatio-temporal data~\cite{lee2023temporal, xu2023pre, li2023videochat, maaz2023video}.

\subsection{Pre-Trained Foundation Models} \label{sec:pre-trained foundation models}
Pre-trained foundation models refer to large-scale pre-trained models that can be adapted to solve a wide range of downstream tasks. In the literature, LLMs and MLLMs can also be viewed as PFMs because they are large-scale pretrained models with broad transferability. To avoid ambiguity, we use the term PFM in the subsequent taxonomy in a narrower operational sense: it denotes pre-trained or foundation models whose backbone, pre-training data, or adaptation objective is designed beyond general language modeling, especially for temporal, spatial, relational, visual, or multimodal data~\cite{zhou2023comprehensive, bommasani2021opportunities}. PFMs constitute a broader category of models that are characterized by their emergent capabilities and homogenization in effectively solving different tasks with the consolidation of methodologies in building AI systems~\cite{bommasani2021opportunities}, which are significantly different from task-specific models. In general, the capabilities of PFMs are in three key dimensions: \textit{modality bridging}, \textit{reasoning and planning}, and \textit{interaction}.

The first aspect refers to multimodal models~\cite{xu2023multimodal}, e.g., MLLMs like vision-language models, which have shown remarkable results in unifying the language and vision modalities~\cite{awais2023foundational}. For example, CLIP~\cite{clip} initially proposed to bridge the gap between images and texts, and SAM~\cite{kirillov2023segment} further extends the concept of textual prompting to visual prompting. Other recent works, such as NExT-GPT~\cite{wu2023next}, further enlarge the boundary and even allow for the bridging of multiple different modalities. It is to be expected that data are naturally multimodal in the real world, such as in clinical medicine~\cite{singhal2023large}, where time series and spatio-temporal data are frequently involved like ECG and medical events. This also motivates recent research interests in multimodal time series and spatio-temporal data~\cite{xue2022promptcast, xu2023mplug}. A brief roadmap of PFMs is given in \shortautoref{fig:roadmap} (\textbf{right}), where VLMs remain the most popular research topic in this area but other PFMs related to time series and spatio-temporal data are emerging, though still in the early stage of development.

The second aspect highlights the reasoning and planning capabilities of PFMs. Typical examples include CoT~\cite{wei2022chain} and GoT~\cite{besta2023graph} in LLMs, as well as task-planning agents~\cite{wu2023embodied, wang2023describe}. The last aspect defines the interaction capabilities of PFMs, and this includes both actioning and communication. 
In this survey, we mainly focus on PFMs for time series and spatio-temporal data, in which most of them remain in the early stage of development and far from reaching the second and third aspects mentioned above. We refer readers to \cite{bommasani2021opportunities} for more details on PFMs.

\subsection{Time Series and Spatio-Temporal Data}\label{sec:time series and spatial-temporal data}
Temporal data, notably time series and spatio-temporal data, serve as foundational data categories for a myriad of real-world applications~\cite{wen2022robust,jin2023spatio}. A time series is commonly defined as an ordered sequence of data points, organized by their occurrence in time. These sequences can be either \textit{univariate} or \textit{multivariate}. For example, daily temperature readings from a city would form a univariate time series, whereas combining daily temperature and humidity data would create a multivariate time series.
In the sequel, we use bold uppercase letters (e.g., $\mathbf{X}$) for matrices, bold lowercase letters (e.g., $\mathbf{x}$) for vectors, calligraphic uppercase letters (e.g., $\mathcal{X}$) for sets, and standard lowercase letters (e.g., $x$) for scalars.
Formally, we adhere to the definitions of time series data outlined in \cite{jin2023survey}, which we summarize below.

\begin{definition}[Time Series Data]
    A univariate time series $\mathbf{x} = \{x_1, x_2, \cdots, x_T\} \in \mathbb{R}^{T}$ is a sequence of $T$ data points indexed in time order, where $x_t \in \mathbb{R}$ is the value of the time series at time $t$. A multivariate time series $\mathbf{X} = \{\mathbf{x}_1, \mathbf{x}_2, \cdots, \mathbf{x}_T\} \in \mathbb{R}^{T \times D}$ is a sequence of $T$ data points indexed in time order but with $D$ dimensions, where $\mathbf{x}_t \in \mathbb{R}^{D} (1 \leq t \leq T)$ denotes the values of the time series at time $t$ along $D$ channels. We refer the reader to \cite{jin2023survey} for more in-depth discussions on this.
\end{definition}

\noindent Spatio-temporal data, on the other hand, are a sequence of data points organized by both temporal and spatial dimensions, encapsulating a wide variety of constructs, including \textit{spatio-temporal graphs} (STGs)~\cite{jin2023survey}, \textit{temporal knowledge graphs} (TKGs)~\cite{ji2021survey}, \textit{videos}~\cite{khan2022transformers}, \textit{point cloud streaming} (PCS)~\cite{xiao2023unsupervised}, \textit{trajectories}~\cite{atluri2018spatio}, among others. In this study, our primary focus is on the first three categories, which are highly representative and align closely with the current research interests. Formally, we follow the definitions of STGs and TKGs in ~\cite{jin2023survey} and \cite{ji2021survey}, which we summarize as follows.

\begin{definition}[Spatio-Temporal Graphs]
    A spatio-temporal graph $\mathcal{G} = \{\mathcal{G}_1, \mathcal{G}_2, \cdots, \mathcal{G}_T\}$ is a sequence of $T$ static graph snapshots indexed in time order, where $\mathcal{G}_t = (\mathcal{V}_t, \mathcal{E}_t)$ denotes a snapshot at time $t$; $\mathcal{V}_t$ and $\mathcal{E}_t$ are sets of nodes and edges at time $t$. The corresponding adjacency and node feature matrices are defined as $\mathbf{A}_t \in \mathbb{R}^{N \times N}$ and $\mathbf{X}_t \in \mathbb{R}^{N \times D}$, where $\mathbf{A}_t = \{a_{ij}^t\}$ and $a_{ij}^t \neq 0$ if there is an edge between node $i$ and $j$, s.t. $e_{ij}^t \in \mathcal{E}_t$. Here, $N=|\mathcal{V}_t|$ is the number of nodes and $D$ is the dimension of node features.
\end{definition}

\begin{definition}[Temporal Knowledge Graphs]
    A temporal knowledge graph $\mathcal{G} = \{\mathcal{G}_1, \mathcal{G}_2, \cdots, \mathcal{G}_T\}$ is a sequence of $T$ static knowledge graph snapshots indexed in time order, where $\mathcal{G}_t = (\mathcal{E}_t, \mathcal{R}_t)$ is a snapshot consisting of the entity and relation sets at time $t$. Specifically, $\mathcal{E}_t$ encapsulates both subject and object entities, and $\mathcal{R}_t$ denotes the set of relations between them. In a temporal knowledge graph, entities and relations may possess different features, denoted by $\mathbf{X}^{e}_t \in \mathbb{R}^{|\mathcal{E}_t| \times D_e}$ and $\mathbf{X}^{r}_t \in \mathbb{R}^{|\mathcal{R}_t| \times D_r}$, where $D_e$ and $D_r$ are feature dimensions.
\end{definition}

\noindent Video data can also be interpreted as a type of spatio-temporal data, which is usually defined as a sequence of images indexed in time order; thus, we define it as follows in a similar manner.

\begin{definition}[Video Data]
    Let $\mathcal{V} = \{\mathbf{F}_1, \mathbf{F}_2, \cdots, \mathbf{F}_T\}$ be a video consisting of $T$ frames indexed in time order, where $\mathbf{F}_t$ denotes the $t$-th frame. For simplicity, we let each frame $\mathbf{F}_t$ be a matrix of pixels, i.e., $\mathbf{F}_t \in \mathbb{R}^{H \times W \times C}$, without considering other metadata, where $H$, $W$, and $C$ are the height, width, and color channels of the frame, respectively.
\end{definition}

\noindent Given the definitions above, we then briefly introduce the representative tasks associated with each data category, which are showcased in \shortautoref{fig:introduction} and summarized below.

\begin{itemize}[leftmargin=*]
    \item \textbf{Time Series Tasks.} This area typically includes four principal analytical tasks: \textit{forecasting}, \textit{classification}, \textit{anomaly detection}, and \textit{imputation}. In forecasting, the goal is to predict the future values of a time series, which can be further categorized into short-term and long-term forecasting depending on the prediction horizon. In classification, the goal is to classify the input time series into different categories. Time series anomaly detection can also be understood as a special type of classification task, where we aim to identify the abnormal time series from the normal ones. In imputation, the goal is to fill in the missing values in time series. Without loss of generality, we refer the reader to \cite{jin2023survey} for more details.
    \item \textbf{Spatio-Temporal Graph Tasks.} The predominant downstream task in spatio-temporal graph is \textit{forecasting}, which aims to predict the future node features by referring to the historical attributive and structural information. Typical examples include traffic forecasting~\cite{jin2023spatio} and some on-demand services~\cite{jin2023survey}. Other common tasks include \textit{link prediction}~\cite{divakaran2020temporal} and \textit{node/graph classification}~\cite{jia2020graphsleepnet}, where the former aims to predict the existence of an edge based on the historical information, and the latter aims to classify the nodes or graphs into different categories. A more in-depth discussion can be found in \cite{sahili2023spatio}.
    \item \textbf{Temporal Knowledge Graph Tasks.} There are two important tasks in temporal knowledge graphs: \textit{completion} and \textit{forecasting}. The former mainly aims to impute missing relations within the graph, while the latter focuses on predicting future relations. For an expanded discussion, please consult \cite{cai2022temporal, han2021temporal}.
    \item \textbf{Video Tasks.} In the field of computer vision, video data encompasses several core tasks such as \textit{detection}, \textit{captioning}, \textit{anticipation}, and \textit{querying}. Detection aims to identify specific objects or actions within the video. Captioning seeks to generate natural language descriptions for the video content. Anticipation involves predicting forthcoming frames in a video sequence. Lastly, querying aims to retrieve video segments pertinent to a specific query. Notably, these tasks often traverse multiple modalities and have received considerable attention compared to the previously mentioned data types. For additional insights, the reader is directed to \cite{awais2023foundational}.
\end{itemize}

%% file: sections/3_taxonomy.tex
\section{Overview and Categorization} \label{sec:overview and categorization}

\input{figures/taxonomy_plot}

In this section, we provide an overview and categorization of the large models for time series and spatio-temporal data. Our survey is structured along four main dimensions: data categories, model architectures, model scopes, and application domains or tasks. A detailed synopsis of the related works can be found in \shortautoref{fig:Taxonomy} and \shortautoref{tab:summary}. We primarily divide the existing body of literature into two major categories: \textit{large models for time series data} (LM4TS) and \textit{large models for spatio-temporal data} (LM4STD).

In the first category, we subdivide the existing literature into two types: \textit{LLMs for time series data} (LLM4TS) and \textit{PFMs for time series data} (PFM4TS). The former refers to leveraging LLMs for solving time series tasks, irrespective of whether LLMs are fine-tuned or frozen during the adaptation process. The latter, on the other hand, focuses on the development of PFMs explicitly designed for various time series tasks. Thus, the LLM/PFM distinction in our taxonomy is practical rather than ontological. LLM4TS and LLM4STD emphasize the repurposing of language-centered foundation models, while PFM4TS and PFM4STD emphasize temporal or spatio-temporal foundation models developed as general backbones for the target data modalities. It is important to note that the field of PFM4TS is relatively nascent; existing models may not fully encapsulate the potential of general-purpose PFMs as defined in \shortautoref{sec:pre-trained foundation models}. Nevertheless, they offer valuable insights into the future development of this field. Thus, we also include them in this survey and categorize them as PFM4TS. For each of these subdivisions, we further classify them into either \textit{general-purpose} or \textit{domain-specific} models, depending on whether the models are designed for solving general time series tasks or restricted to specific domains, including but not limited to transportation, finance, and healthcare.

In the category of LM4STD, we employ a similar taxonomy, defining \textit{LLMs for spatio-temporal data} (LLM4STD) and \textit{PFMs for spatio-temporal data} (PFM4STD). Unlike time series data, spatio-temporal data encompass a broader array of entities spanning multiple domains; thus we explicitly categorize LLM4STD and PFM4STD by their relevant domains. While we offer a categorization based on model scopes in \shortautoref{tab:summary} to facilitate the understanding of the existing literature, this categorization is conditioned on different data modalities (e.g., STGs and videos), as they typically possess different problem definitions. Therefore, this classification is omitted in \shortautoref{fig:Taxonomy} for the sake of brevity. Herein, we focus on the three most prominent domains/modalities, using them as subcategories: spatio-temporal graphs, temporal knowledge graphs, and video data. For each of them, we summarize the representative tasks as the leaf nodes (see \shortautoref{fig:Taxonomy}), which is similar to LM4TS. Notably, PFM4STD has seen more extensive development than its time series counterpart. Current research is mainly geared towards STGs and video data, often featuring enhanced capabilities of PFMs such as multimodal bridging and reasoning.

With this foundational understanding of data categories, model architectures, model scopes, and application domains/tasks, we will delve into the specifics of LM4TS and LM4STD in the subsequent sections, labeled \shortautoref{sec: Large Models for Time Series Data} and \shortautoref{sec: Large Models for Spatio-Temporal Data}, respectively. Within these sections, we will review pivotal contributions in LLM4TS and PFM4TS as outlined in \shortautoref{sec: LLMs in Time Series} and \shortautoref{sec: PFMs in Time Series}. Similarly, in \shortautoref{sec: spatio-temporal graphs}, we will summarize key research related to LLMs and PFMs for STGs, while in \shortautoref{sec: temporal knowledge graphs} and \shortautoref{sec: video data}, we will provide an overview of pertinent literature concerning LLMs and PFMs for TKGs and Video Data.

\input{tables/summary_of_literature}

%% file: figures/taxonomy_plot.tex
\tikzstyle{my-box}=[
    rectangle,
    draw=hidden-draw,
    rounded corners,
    text opacity=1,
    minimum height=1.5em,
    minimum width=5em,
    inner sep=2pt,
    align=center,
    fill opacity=.5,
    line width=0.8pt,
]
\tikzstyle{leaf}=[my-box, minimum height=1.5em,
    fill=hidden-pink!80, text=black, align=left, font=\normalsize,
    inner xsep=2pt,
    inner ysep=4pt,
    line width=0.8pt,
]
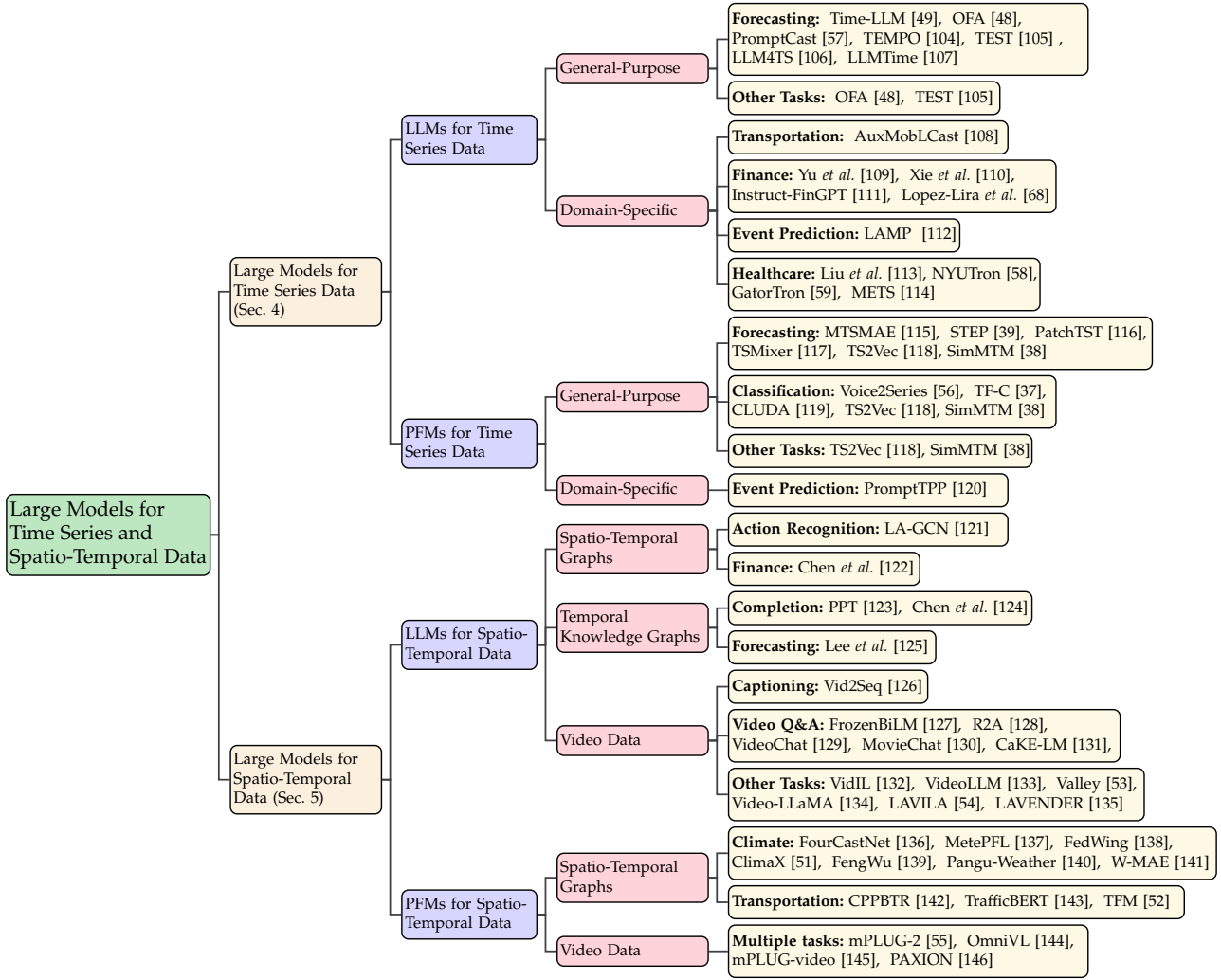
\begin{figure*}[t]
    \centering
    \resizebox{0.98\textwidth}{!}{
        \begin{forest}
            forked edges,
            for tree={
                fill=level0!80,
                grow=east,
                reversed=true,
                anchor=base west,
                parent anchor=east,
                child anchor=west,
                base=left,
                font=\large,
                rectangle,
                draw=hidden-draw,
                rounded corners,
                align=left,
                minimum width=4em,
                edge+={darkgray, line width=1pt},
                s sep=3pt,
                inner xsep=2pt,
                inner ysep=3pt,
                line width=0.8pt,
                ver/.style={rotate=90, child anchor=north, parent anchor=south, anchor=center},
            },
            where level=1{text width=9em,font=\normalsize,fill=level1!80,}{},
            where level=2{text width=8em,font=\normalsize,fill=level2!80,}{},
            where level=3{text width=9em,font=\normalsize,fill=level3!60,}{},
            where level=4{text width=5em,font=\normalsize,}{},
            [
                Large Models for \\ Time Series and \\ Spatio-Temporal Data
                [
                    Large Models for \\ Time Series Data  \\ (\shortautoref{sec: Large Models for Time Series Data})
                    [
                        LLMs for Time \\ Series Data
                        [
                            General-Purpose 
                            [ 
                                \textbf{Forecasting: } LLM4TS~\cite{chang2023llm4ts}{, } TimeCMA~\cite{liu2025timecma}{, }  
                                Time-VLM~\cite{zhong2025time}{, } \\
                                Time-LLM~\cite{jin2023timellm}{, }
                                TEMPO~\cite{cao2023tempo}{, }
                                UniTime~\cite{liu2024unitime}{, }  
                                ChatTime~\cite{wang2025chattime}{, } \\
                                LLMTime~\cite{gruver2023large}{, } PromptCast~\cite{xue2022promptcast}
                                , leaf, text width=28em
                            ]
                            [ 
                                \textbf{Multiple Tasks: }
                                Time-MQA~\cite{kong2025time}{, }
                                TimeOmni-1~\cite{guan2025timeomni}{, } \\
                                ChatTS~\cite{xie2024chatts}{, }
                                Time-MMD~\cite{liu2024time}
                                , leaf, text width=24em
                            ]
                        ]
                        [
                            Domain-Specific
                            [   
                                \textbf{Transportation: } 
                                Traffic-R1~\cite{zou2025traffic}{, }
                                LEAF~\cite{zhao2025embracing}{, } \\
                                LLM-Mob~\cite{wang2023i}{, } 
                                xTP-LLM~\cite{guo2024towards}
                                , leaf, text width=21em
                            ]
                            [   
                                \textbf{Finance: }
                                M2VN~\cite{kong2025fusing}{, }
                                DINN~\cite{hwang2025decision}{, }
                                GPT4FTS~\cite{jia2025beyond}{, } \\
                                Lopez-Lira \textit{et al.}~\cite{lopez2023can}
                                , leaf, text width=23em
                            ]
                            [   
                                \textbf{Healthcare: } 
                                MedualTime~\cite{ye2024medualtime}{, }
                                OpenTSLM~\cite{langer2025opentslm}{, } \\
                                MedTsLLM~\cite{chan2024leveraging}{, } 
                                NYUTron~\cite{jiang2023health}
                                , leaf, text width=22em
                            ]
                        ]
                    ]
                    [
                        PFMs for Time \\ Series Data
                        [
                            General-Purpose
                            [
                                \textbf{Forecasting: } 
                                Chronos-2~\cite{ansari2025chronos}{, }
                                Time-MoE~\cite{shi2024time}{, }
                                Chronos~\cite{ansari2024chronos}{, } \\
                                Sundial~\cite{liu2025sundial}{, } 
                                Time-FFM~\cite{liu2024time-ffm}{, }
                                Lag-llama~\cite{rasul2023lag}{, } 
                                TimesFM~\cite{das2024decoder}{, } \\
                                MOIRAI~\cite{woo2024unified}{, } 
                                TimeMixer++~\cite{wang2024timemixer++}{, } 
                                PatchTST~\cite{nie2022time}
                                , leaf, text width=28em
                            ]
                            [
                                \textbf{Multiple Tasks: } 
                                Timer~\cite{liu2024timer}{, }
                                MOMENT~\cite{goswami2024moment}
                                , leaf, text width=21em   
                            ]
                        ]
                        [
                            Domain-Specific
                            [
                                \textbf{Finance: } 
                                SSPT~\cite{wang2025pre}{, }
                                DELPHYNE~\cite{ding2025delphyne}
                                , leaf, text width=18em
                            ]
                            [
                                \textbf{Healthcare: } 
                                ECG-FM~\cite{mckeen2024ecg}{, }
                                MIRA~\cite{li2025mira}{, }
                                Mathew \textit{et al.}~\cite{mathew2024foundation}
                                , leaf, text width=28em
                            ]
                        ]
                    ]
                ]
                [                
                    Large Models for \\ Spatio-Temporal \\ Data (\shortautoref{sec: Large Models for Spatio-Temporal Data})
                    [
                        LLMs for Spatio-\\Temporal Data
                        [
                            Spatio-Temporal \\Graphs
                            [
                                \textbf{Forecasting: } 
                                FSTLLM~\cite{jiangfstllm}{, }
                                ST-LINK~\cite{jeon2025st}{, } \\
                                RePST~\cite{wang2025repst}{, }
                                STG-LLM~\cite{liu2024can}
                                , leaf, text width=19em
                            ]
                            [
                                \textbf{Transportation: }  
                                GSF-LLM~\cite{wang2025gsf}{, }
                                ST-LLM+~\cite{liu2025st}{, } \\
                                LLM-TFP~\cite{cheng2025llm}{, }
                                UrbanGPT~\cite{urbangpt}
                                , leaf, text width=22em
                            ]
                        ]
                        [
                            Temporal \\Knowledge Graphs
                            [
                                \textbf{Completion: } 
                                Ong \textit{et al.}~\cite{ong2025dynamic}{, } 
                                TeRDy~\cite{liu2025terdy}{, } 
                                Luo \textit{et al.}~\cite{luo2024chain}
                                , leaf, text width=27em
                            ]
                            [
                                \textbf{Forecasting: } 
                                G2S~\cite{bai2025g2s}{, }
                                AnRe~\cite{tang2025anre}{, }
                                Xia \textit{et al.}~\cite{xia2024chain}{, } \\
                                LLM-DA~\cite{wang2024large}{, } 
                                LATE~\cite{zhang2024harnessing}
                                , leaf, text width=23em
                            ]
                        ]
                        [
                            Video Data
                            [
                                \textbf{Video Q\&A: } 
                                MovieChat~\cite{song2023moviechat}{, }
                                CaKE-LM~\cite{su2023language}
                                , leaf, text width=21em
                            ]
                            [
                                \textbf{Sport: } 
                                Jiang \textit{et al.}~\cite{jiang2025domain}{, }
                                MatchVision~\cite{rao2025towards}
                                , leaf, text width=19em
                            ]
                            [
                                \textbf{Traffic: } 
                                TRIVIA~\cite{qasemi2023traffic}
                                , leaf, text width=10.5em
                            ]
                            [
                                \textbf{Multiple Tasks: } 
                                Video-LLaMA-3~\cite{zhang2025videollama}{, }
                                Valley~\cite{luo2023valley}{, }
                                MA-LMM~\cite{he2024ma}{, } \\
                                Video-XL~\cite{shu2025video}{, }
                                video-SALMONN-o1~\cite{sun2025video}{, }
                                BOLT~\cite{liu2025bolt}{, }
                                LAVENDER~\cite{li2023lavender} \\
                                Video-ChatGPT~\cite{maaz2023video}{, }
                                ProgGen~\cite{tang2025programmatic}{, }
                                Video-LLaMA~\cite{zhang2023video}{, }
                                Apollo~\cite{zohar2025apollo}
                                , leaf, text width=33em
                            ]
                        ]
                    ]
                    [
                        PFMs for Spatio-\\Temporal Data
                        [
                            Spatio-Temporal \\Graphs
                            [
                                \textbf{Climate: } 
                                FourCastNet-3~\cite{bonev2025fourcastnet}{, }
                                ClimaX~\cite{nguyen2023climax}{, }
                                Pangu-Weather~\cite{bi2022pangu}
                                , leaf, text width=29em
                            ]
                            [
                                \textbf{Transportation: } 
                                TrafficBERT~\cite{jin2021trafficbert}{, }
                                TFM~\cite{wang2023building}
                                , leaf, text width=22em
                            ]
                            [
                                \textbf{Multiple Tasks: } 
                                UMSST~\cite{zhong5297158unified}{, }
                                UniSTD~\cite{tang2025unistd}
                                , leaf, text width=22em
                            ]
                        ]
                        [
                            Video Data
                            [
                                \textbf{Multiple Tasks: } 
                                MSTA~\cite{chen2025efficient}{, }
                                InternVideo~\cite{wang2024internvideo2}{, } \\
                                mPLUG-Owl2~\cite{ye2024mplug}{, }
                                mPLUG-2~\cite{xu2023mplug}{, }
                                OmniVL~\cite{wang2022omnivl}{, } \\
                                mPLUG-video~\cite{xu2023youku}
                                , leaf, text width=24em
                            ]
                        ]
                      ]
                    ]
                ]
            ]
        \end{forest}
    }
\caption{A comprehensive taxonomy of large models for time series and spatio-temporal data from the perspectives of methodologies (i.e., LLMs vs. PFMs), motivations (e.g., general vs. domain-specific purposes), and applications.}
\vspace{-5mm}
\label{fig:Taxonomy}
\end{figure*}

%% file: tables/summary_of_literature.tex
\begin{table*}[htbp]
    \footnotesize
    \centering
    \caption{Summary of LLMs and PFMs for time series and spatio-temporal data modeling. General-purpose methods without a specific task/domain are marked as ``-''. The table complements Fig.~\ref{fig:Taxonomy} with bibliographic and application-level details.}
    \vspace{-2mm}
    \resizebox{\textwidth}{!}{
    \begin{tabular}{c|cccccccccc}
    \toprule\toprule
    \textbf{} & \textbf{Category} & \textbf{Method} & \textbf{Scope} & \textbf{Task / Domain} & \textbf{Institute} & \textbf{Venue} & \textbf{Year} \\
    \hline
    \multirow{42}{*}{\rotatebox[origin=c]{90}{\textbf{Large Models for Time Series Data}}}
    & \multirow{25}{*}{\rotatebox[origin=c]{0}{LLMs for Time Series Data}} 
    &  
    LLM4TS~\cite{chang2023llm4ts} & General-Purposed & Forecasting & NYCU & ACM TIST & 2025 \\
    & & 
    TimeCMA~\cite{liu2025timecma} & General-Purposed & Forecasting & NTU & AAAI & 2025 \\
    & & 
    Time-VLM~\cite{zhong2025time} & General-Purposed & Forecasting & HKUST & ICML & 2025 \\
    & & 
    Time-MQA~\cite{kong2025time} & General-Purposed & - & Oxford & ACL & 2025 \\
    & & 
    TimeOmni-1~\cite{guan2025timeomni} & General-Purposed & - & Griffith University & ICLR & 2026 \\
    & & 
    ChatTS~\cite{xie2024chatts} & General-Purposed & - & Tsinghua & VLDB & 2025 \\
    & & 
    Time-MMD~\cite{liu2024time} & General-Purposed & - & Georgia Tech & NeurIPS & 2024 \\
    & & Time-LLM~\cite{jin2023timellm} & General-Purposed & Forecasting & Monash/Alibaba/Ant & ICLR & 2024 \\
    & & 
    TEMPO~\cite{cao2023tempo} & General-Purposed & Forecasting & USC/Google & ICLR & 2024 \\
    & &
    UniTime~\cite{liu2024unitime} & General-Purposed & Forecasting & NUS & WWW & 2024 \\
    & &
    ChatTime~\cite{wang2025chattime} & General-Purposed & Forecasting & BUPT & AAAI & 2025 \\
    & & 
    LLMTime~\cite{gruver2023large} & General-Purposed & Forecasting & NYU/CMU & NeurIPS & 2023 \\
    & & 
    PromptCast~\cite{xue2022promptcast} & General-Purposed & Forecasting & UNSW & IEEE Trans. Knowl. Data Eng. & 2024 \\
    & & 
    MedualTime~\cite{ye2024medualtime} & Domain-Specific & Healthcare & HKUST & IJCAI & 2025 \\
    & & 
    Traffic-R1~\cite{zou2025traffic} & Domain-Specific & Transportation & HKUST & arXiv & 2025 \\
    & & 
    LEAF~\cite{zhao2025embracing} & Domain-Specific & Transportation & PKU & ACL & 2025 \\
    & & 
    M2VN~\cite{kong2025fusing} & Domain-Specific & Finance & Oxford & ICAIF & 2025 \\
    & & 
    DINN~\cite{hwang2025decision} & Domain-Specific & Finance & Oxford & arXiv & 2025 \\
    & & 
    GPT4FTS~\cite{jia2025beyond} & Domain-Specific & Finance & Tongji University & arXiv & 2026 \\
    & & 
    OpenTSLM~\cite{langer2025opentslm} & Domain-Specific & Healthcare & Stanford & arXiv & 2026 \\
    & & 
    Lopez-Lira \textit{et al.}~\cite{lopez2023can} & Domain-Specific & Finance & UFL & arXiv & 2025 \\
    & & 
    LLM-Mob~\cite{wang2023i} & Domain-Specific & Transportation & UCL/Liverpool & arXiv & 2024 \\
    & & 
    xTP-LLM~\cite{guo2024towards} & Domain-Specific & Transportation & HKUST & Commun.Transp.Res. & 2024 \\
    & & 
    MedTsLLM~\cite{chan2024leveraging} & Domain-Specific & Healthcare & JHU & MLHC & 2024 \\ 
    & & 
    NYUTron~\cite{jiang2023health} & Domain-Specific & Healthcare & NYU Langone Health & Nature & 2023 \\
    \cline{2-8}
    & \multirow{17}{*}{\rotatebox[origin=c]{0}{PFMs for Time Series Data}} 
    & 
    Chronos-2~\cite{ansari2025chronos} & General-Purposed & Forecasting & Amazon & arXiv & 2025 \\
    & & 
    Time-MoE~\cite{shi2024time} & General-Purposed & Forecasting & Xiaohongshu/Princeton & ICLR & 2025 \\
    & & 
    TimeMixer++~\cite{wang2024timemixer++} & General-Purposed & Forecasting & Griffith/HKUST & ICLR & 2025 \\
    & & 
    Sundial~\cite{liu2025sundial} & General-Purposed & Forecasting & Tsinghua & ICML & 2025 \\
    & & 
    Time-FFM~\cite{liu2024time-ffm} & General-Purposed & Forecasting & HKUST & NeurIPS & 2024 \\
    & & 
    Lag-llama~\cite{rasul2023lag} & General-Purposed & Forecasting & Morgan Stanley & NeurIPS Workshop & 2023 \\
    & & 
    TimesFM~\cite{das2024decoder} & General-Purposed & Forecasting & Google & ICML & 2024 \\
    & & 
    Timer~\cite{liu2024timer} & General-Purposed & - & Tsinghua & ICML & 2024 \\
    & & 
    MOIRAI~\cite{woo2024unified} & General-Purposed & Forecasting & Salesforce & ICML & 2024 \\
    & & 
    MOMENT~\cite{goswami2024moment} & General-Purposed & - & CMU & ICML & 2024 \\
    & & 
    Chronos~\cite{ansari2024chronos} & General-Purposed & Forecasting & Amazon & TMLR & 2024 \\
    & & 
    PatchTST~\cite{nie2022time} & General-Purposed & Forecasting & Princeton/IBM & ICLR & 2023 \\
    & & 
    SSPT~\cite{wang2025pre} & Domain-Specific & Finance & Edinburgh & KDD & 2025 \\
    & & 
    DELPHYNE~\cite{ding2025delphyne} & Domain-Specific & Finance & CMU/Bloomberg & arXiv & 2025 \\
    & & 
    ECG-FM~\cite{mckeen2024ecg} & Domain-Specific & Healthcare & UofT & JAMIA Open & 2025 \\
    & & 
    MIRA~\cite{li2025mira} & Domain-Specific & Healthcare & Microsoft & NeurIPS & 2025 \\
    & & 
    Mathew \textit{et al.}~\cite{mathew2024foundation} & Domain-Specific & Healthcare & Eko Health & npj Cardiovasc. Health. & 2024 \\

    \hline\hline
    \multirow{45}{*}{\rotatebox[origin=c]{90}{\textbf{Large Models for Spatio-Temporal Data}}}
    & \multirow{8}{*}{\rotatebox[origin=c]{0}{LLMs for Spatio-Temporal Graphs}}
    & 
    FSTLLM~\cite{jiangfstllm} & General-Purposed & Forecasting & NTU & ICML & 2025 \\
    & & 
    ST-LINK~\cite{jeon2025st} & General-Purposed & Forecasting & POSTECH & CIKM & 2025 \\
    & &
    RePST~\cite{wang2025repst} & General-Purposed & Forecasting & HKUST & IJCAI & 2025 \\
    & &
    STG-LLM~\cite{liu2024can} & General-Purposed & Forecasting & DLUT & arXiv & 2024 \\
    & & 
    GSF-LLM~\cite{wang2025gsf} & Domain-Specific & Transportation & Tongji University & Sensors & 2025 \\
    & & 
    ST-LLM+~\cite{liu2025st} & Domain-Specific & Transportation & NTU & IEEE Trans. Knowl. Data Eng. & 2025 \\
    & & 
    LLM-TFP~\cite{cheng2025llm} & Domain-Specific & Transportation & NUPT & Appl. Soft Comput. & 2025 \\
    & & 
    UrbanGPT~\cite{urbangpt} & Domain-Specific & Transportation & HKU & KDD & 2024 \\
    \cline{2-8}
    & \multirow{7}{*}{\rotatebox[origin=c]{0}{PFMs for Spatio-Temporal Graphs}}
    & 
    UMSST~\cite{zhong5297158unified} & General-Purposed & - & UTS & Knowl. -Based Syst. & 2025 \\
    & & 
    UniSTD~\cite{tang2025unistd} & General-Purposed & - & CUHK & CVPR & 2025 \\
    & & 
    FourCastNet-3~\cite{bonev2025fourcastnet} & Domain-Specific & Climate & NVIDIA & arXiv & 2025 \\
    & & 
    ClimaX~\cite{nguyen2023climax} & Domain-Specific & Climate & UCLA/Microsoft & ICML & 2023 \\
    & & 
    Pangu-Weather~\cite{bi2022pangu} & Domain-Specific & Climate & Huawei & Nature & 2023 \\
    & & 
    TFM~\cite{wang2023building} & Domain-Specific & Transportation & Shanghai AI Lab & IEEE ITSC & 2023 \\
    & & 
    TrafficBERT~\cite{jin2021trafficbert} & Domain-Specific & Transportation & CAU & Expert Syst. Appl. & 2021 \\
    
    \cline{2-8}
    & \multirow{8}{*}{\rotatebox[origin=c]{0}{LLMs for Temporal Knowledge Graphs}}
    & 
    Ong \textit{et al.}~\cite{ong2025dynamic} & General-Purposed & Completion & IC & ESWA & 2025 \\
    & & 
    TeRDy~\cite{liu2025terdy} & General-Purposed & Completion & Tsinghua & ACL & 2025 \\
    & & 
    G2S~\cite{bai2025g2s} & General-Purposed & Forecasting & CAS & ACL & 2025 \\
    & & 
    AnRe~\cite{tang2025anre} & General-Purposed & Forecasting & HIT & ACL & 2025 \\
    & & 
    Luo \textit{et al.}~\cite{luo2024chain} & General-Purposed & Completion & Tsinghua & arXiv & 2024 \\
    & & 
    Xia \textit{et al.}~\cite{xia2024chain} & General-Purposed & Forecasting & CAS & ACL & 2024 \\
    & & 
    LLM-DA~\cite{wang2024large} & General-Purposed & Forecasting & BJUT & NeurIPS & 2024 \\
    & & 
    LATE~\cite{zhang2024harnessing} & General-Purposed & Forecasting & USTC & ICAIRC & 2024 \\

    \cline{2-8}
    & \multirow{16}{*}{\rotatebox[origin=c]{0}{LLMs for Video Data}}
    & 
    Video-LLaMA-3~\cite{zhang2025videollama} & General-Purposed & - & Alibaba & arXiv & 2025 \\
    & & 
    Valley~\cite{luo2023valley} & General-Purposed & - & ByteDance/Fudan & arXiv & 2025 \\ 
    & & 
    Apollo~\cite{zohar2025apollo} & General-Purposed & - & Meta & CVPR & 2025 \\
    & &
    Video-XL~\cite{shu2025video} & General-Purposed & - & SJTU & CVPR & 2025 \\
    & &
    video-SALMONN-o1~\cite{sun2025video} & General-Purposed & - & Cambridge/ByteDance & ICML & 2025 \\
    & &
    BOLT~\cite{liu2025bolt} & General-Purposed & - & KAUST & CVPR & 2025 \\
    & &
    ProgGen~\cite{tang2025programmatic} & General-Purposed & - & Cornell & ICLR Workshop & 2025 \\
    & &
    MovieChat~\cite{song2023moviechat} & General-Purposed & Video Question Answering & ZJU & CVPR & 2024 \\
    & & 
    Video-ChatGPT~\cite{maaz2023video} & General-Purposed & - & MBZUAI & ACL & 2024 \\
    & & 
    MA-LMM~\cite{he2024ma} & General-Purposed & - & UMD/Meta & CVPR & 2024 \\
    & & 
    CaKE-LM~\cite{su2023language} & General-Purposed & Video Question Answering & NTU & CVPR & 2023 \\
    & & 
    Video-LLaMA~\cite{zhang2023video} & General-Purposed & - & Alibaba & EMNLP & 2023 \\
    & & 
    LAVENDER~\cite{li2023lavender} & General-Purposed & - & Microsoft & CVPR & 2023 \\
    & & 
    Jiang \textit{et al.}~\cite{jiang2025domain} & Domain-Specific & Sport-Domain & MIT/Amazon & CVPR & 2025 \\
    & & 
    MatchVision~\cite{rao2025towards} & Domain-Specific & Sport-Domain & SJTU/Alibaba & CVPR & 2025 \\
    & & 
    TRIVIA~\cite{qasemi2023traffic} & Domain-Specific & Traffic-Domain VQA & USC & IEEE ITSC & 2023 \\

    \cline{2-8}
    & \multirow{6}{*}{\rotatebox[origin=c]{0}{PFMs for Video Data}}
    & 
    MSTA~\cite{chen2025efficient} & General-Purposed & - & Ant & CVPR & 2025 \\
    & &
    InternVideo~\cite{wang2024internvideo2} & General-Purposed & - & Shanghai AI Lab & ECCV & 2024 \\
    & &
    mPLUG-Owl2~\cite{ye2024mplug} & General-Purposed & - & Alibaba & CVPR & 2024 \\
    & &
    mPLUG-2~\cite{xu2023mplug} & General-Purposed & - & Alibaba & ICML & 2023 \\
    & & mPLUG-video~\cite{xu2023youku} & General-Purposed & - & Alibaba & arXiv & 2023 \\
    & &
    OmniVL~\cite{wang2022omnivl} & General-Purposed & - & Fudan & NeurIPS & 2022 \\
    \bottomrule\bottomrule
    \end{tabular}
    }
    \vspace{1mm}
    \label{tab:summary}
\end{table*}

%% file: sections/4_lm4ts.tex
\section{Large Models for Time Series Data} \label{sec: Large Models for Time Series Data}
In this section, we delve into the advancements of large language models and pre-trained foundation models in time series analysis. Within each subsection, we further segment our discussion based on whether the models serve general or domain-specific purposes, as outlined in \shortautoref{fig:Taxonomy}.

\subsection{Large Language Models in Time Series} \label{sec: LLMs in Time Series}
Time series analysis is a fundamental problem in many real-world applications, such as sales forecasting, imputation of missing economic time series data, anomaly detection for industrial maintenance, and classification of time series from various domains. Witnessing growth of LLMs in the NLP field, we are particularly interested in exploring the research question of whether we can take advantage of these models and adapt them for time series analysis. In this subsection, we discuss LLM4TS in terms of general-purpose and domain-specific models in the following parts, respectively.

\subsubsection{General Models}
Early general LLM4TS studies mainly examine whether existing LLMs can be adapted to time series through prompting, tokenization, decomposition, or reprogramming. As one of the first efforts in approaching general time series forecasting from an LLM perspective, \cite{xue2022promptcast} formally introduces a novel task: prompt-based time series forecasting --- PromptCast. With the input and output being natural language sentences, PromptCast presents a novel ``code less'' solution for time series forecasting, which could provide a new perspective rather than purely focusing on designing complicated architectures. An instruction dataset (PISA) for the newly introduced task has been released. Another closely related study, LLMTime~\cite{gruver2023large}, demonstrates that LLMs are effective zero-shot time series learners with properly configured tokenization on time series data. TEMPO~\cite{cao2023tempo} remains focused on time series forecasting but incorporates additional fine-grained designs such as time series decomposition and soft prompts. Different from the above methods, a recent work Time-LLM~\cite{jin2023timellm} is proposed to reprogram time series with the source data modality along with natural language-based prompting to unleash the potential of LLMs as effective time series machines, which achieves state-of-the-art performance in various forecasting scenarios, as well as excels in both few-shot and zero-shot settings. UniTime~\cite{liu2024unitime} follows similar concepts, albeit with slightly different technical configurations. LLM4TS~\cite{chang2023llm4ts} further leverages a two-stage recipe — first aligning an LLM to time series data via supervised fine-tuning, then fine-tuning for downstream forecasting. A later trend moves beyond forecasting-oriented adaptation toward multimodal alignment, conversational interaction, and time series reasoning. Recently, ChatTime~\cite{wang2025chattime} and TimeCMA~\cite{liu2025timecma} extend this line by bridging multimodal and cross-modal aspects: ChatTime unifies numerical and textual modalities for joint understanding and zero-shot forecasting, while TimeCMA aligns temporal and linguistic features to enhance accuracy and interpretability. Complementing these model-centric efforts, the Time-MMD~\cite{liu2024time} study supplies a large-scale multi-domain multimodal benchmark that tightly aligns text and series, enabling more rigorous evaluation of multimodal TS methods. Building on multimodal cues, Time-VLM~\cite{zhong2025time} integrates vision–language priors to augment forecasting. Beyond forecasting toward understanding and reasoning, ChatTS~\cite{xie2024chatts} treats multivariate time series as a native modality for conversational Multimodal LLMs using synthetic curricula, and Time-MQA~\cite{kong2025time} unifies multi-task question answering over time series, supporting both numerical analytics and open-ended reasoning. Pushing further, TimeOmni-1~\cite{guan2025timeomni} introduces a time series reasoning suite and a unified model spanning perception, extrapolation, and decision-making, signaling a shift from pattern matching to general time series reasoning.

\subsubsection{Domain-Specific Models} 

\paragraph*{\textbf{Transportation.}}
Recent work in Intelligent Transportation Systems (ITS) has increasingly leveraged LLMs to unify traffic forecasting, multimodal data integration, and interpretability. ~\cite{guo2024towards} introduce xTP-LLM, which transforms heterogeneous traffic features — historical flow, temporal indicators, weather, and POI context — into structured textual prompts, enabling an LLM to jointly predict traffic flow and produce natural-language explanations tied to observed patterns. Domain-specific transportation LLMs have since expanded this direction: Traffic-R1~\cite{zou2025traffic} uses instruction tuning on traffic operation logs to support expert-style reasoning for routing and congestion analysis; LEAF~\cite{zhao2025embracing} combines LLMs with spatial–temporal graph structures to capture network topology and road semantics; and LLM-Mob~\cite{wang2023i} integrates LLMs with trajectory data to support mobility-behavior analysis and scenario reasoning. Together, these models illustrate a growing trend toward LLMs that not only forecast traffic conditions but also interpret, contextualize, and interact with complex transportation data.

\paragraph*{\textbf{Finance.}} Recent work has begun to explore LLM-based architectures in core financial prediction problems. M2VN~\cite{kong2025fusing} proposes a multi-modal volatility forecasting network that combines price, volume, and LLM-derived news embeddings, showing that jointly modeling numerical and textual signals improves equity-market volatility prediction. DINN~\cite{hwang2025decision} introduces a decision-informed neural network for portfolio optimization that integrates LLM-based semantic embeddings and a pretrained language model into a unified forecasting-and-allocation pipeline. GPT4FTS~\cite{jia2025beyond} explicitly enhances an LLM backbone for financial time series prediction via adaptive patch segmentation and dynamic wavelet-style modules, achieving state-of-the-art stock return forecasting performance. Complementing these architecture-oriented contributions, Lopez-Lira et al.~\cite{lopez2023can} use general-purpose LLMs (e.g., ChatGPT) to score news headlines and demonstrate robust return predictability, suggesting that even non–domain-specific models can extract economically meaningful signals from financial text. Collectively, these studies illustrate a shift from purely tabular or classical NLP pipelines toward LLM-centric frameworks that fuse textual and numerical information for tasks such as volatility forecasting, portfolio construction, and return prediction.

\paragraph*{\textbf{Healthcare.}} This is one of the most important and applicable domains of event sequence. By forecasting clinical events, i.e., patients' sequences of time-stamped hospital visits with their symptoms, the clinical models can assist physicians and administrators in making decisions in everyday practice. However, existing structured data-based clinical models have limited use owing to the complexity of data processing, as well as model development and deployment. To address these limitations, \cite{jiang2023health} trains a clinical LLM (NYUTron) based on NYU Langone EHRs and subsequently fine-tunes it across a wide range of clinical and operational predictive tasks. The results show that NYUTron brings significant improvement over traditional models, demonstrating the potential for using clinical LLMs in medicine to read alongside physicians and provide guidance at the point of care. Extending this progress, recent multimodal frameworks such as MedualTime~\cite{ye2024medualtime} and MedTsLLM~\cite{chan2024leveraging} further integrate medical time series with clinical text through dual-adapter and cross-modal learning mechanisms, enabling richer temporal–contextual representations and enhancing predictive performance across diverse clinical scenarios. In parallel, OpenTSLM~\cite{langer2025opentslm} introduces a family of time series language models that treat time series as a native modality alongside text, enabling clinicians and patients to query, interpret, and reason about longitudinal health data directly in natural language. 

\subsection{Pre-Trained Foundation Models in Time Series} \label{sec: PFMs in Time Series}
Beyond leveraging LLMs for time series analysis, pre-trained foundation models for time series (PFM4TS) aim to capture generic temporal patterns that transfer across domains and tasks. In this subsection, we explore PFM4TS, distinguishing between general-purpose and domain-specific models, in parallel with the categorization in \shortautoref{sec: LLMs in Time Series}.

\subsubsection{General Models}
General PFM4TS research is characterized by a shift from task-specific forecasting architectures toward scalable pre-training backbones that transfer across datasets, frequencies, variables, and downstream tasks. Since 2023, research on large-scale, general-purpose time series foundation models has accelerated, inspired by advances in language and vision foundation models. PatchTST~\cite{nie2022time} is an early representative approach: it segments multivariate series into sub-series ``patch'' tokens and applies a channel-independent Transformer backbone, substantially improving long-term forecasting and motivating many subsequent architectures. Building on this foundation, Chronos~\cite{ansari2024chronos} conceptualizes time series as a ``language,'' using large-scale generative pre-training to enable universal zero-shot forecasting across diverse datasets, while Chronos-2~\cite{ansari2025chronos} extends this paradigm from univariate to universal settings, supporting univariate, multivariate, and covariate-informed forecasting within a single encoder-only architecture and achieving state-of-the-art zero-shot performance on recent benchmarks. Parallel efforts explore alternative backbones and objectives: Lag-llama~\cite{rasul2023lag} focuses on general-purpose probabilistic univariate forecasting; TimesFM~\cite{das2024decoder} develops a decoder-only Transformer pre-trained on billions of time points for zero-shot forecasting across domains; MOMENT~\cite{goswami2024moment} provides an open family of models for general-purpose time series analysis beyond forecasting; and Timer~\cite{liu2024timer} casts forecasting, imputation, and anomaly detection into a unified generative pre-training objective.

More recent work further broadens the PFM4TS design space along three axes: data scale, architectural efficiency, and task coverage. MOIRAI~\cite{woo2024unified} proposes a masked encoder-based universal forecasting Transformer trained on the LOTSA corpus, enabling any-variate, multi-frequency zero-shot forecasting with exogenous covariates. Time-FFM~\cite{liu2024time-ffm} instead leverages pretrained language models and federated learning, mapping time series to text tokens and learning domain-adaptive prompts while respecting data-locality constraints. TimeMixer++~\cite{wang2024timemixer++} introduces a fully MLP-based architecture that converts time series into multi-resolution ``time images'' to capture multi-scale, multi-periodic patterns and support a broad range of predictive tasks (forecasting, classification, anomaly detection, imputation). Sundial~\cite{liu2025sundial} advances a generative foundation-model family trained on the trillion-point TimeBench corpus, using continuous patching and a flow-matching-based TimeFlow loss to support flexible point and probabilistic zero-shot forecasting. Finally, Time-MoE~\cite{shi2024time} pushes scaling via a sparse decoder-only mixture-of-experts design trained on the Time-300B dataset, reaching billion-parameter models that activate only a subset of experts per query. Together, these models illustrate a rapid convergence toward scalable, general-purpose PFM4TS with strong zero-shot performance, support for multivariate and covariate-informed inputs, and increasing coverage of broader time series tasks.

\subsubsection{Domain-Specific Models}
Beyond general-purpose PFMs, several models are tailored to specific domains such as finance. SSPT~\cite{wang2025pre} proposes a customized pre-training strategy for stock time series using a dual-layer Transformer with auxiliary objectives (e.g., stock-code and sector prediction, moving-average forecasting) to capture both global temporal patterns and stock-specific structure, improving robustness in non-stationary markets. DELPHYNE~\cite{ding2025delphyne} further targets financial applications by pre-training on a mixture of general and finance-specific time series, reducing negative transfer and achieving strong performance on diverse downstream forecasting tasks.

In healthcare, ECG-FM~\cite{mckeen2024ecg} is an open-weight Transformer-based foundation model pre-trained on millions of electrocardiogram (ECG) recordings with ECG-specific augmentations, contrastive learning, and masking objectives, yielding embeddings that transfer well to diverse diagnostic tasks (e.g., arrhythmia and ventricular dysfunction). MIRA~\cite{li2025mira} extends this idea to heterogeneous clinical time series via continuous-time positional encodings and mixture-of-experts layers, enabling zero-shot and fine-tuned forecasting across institutions. Mathew et al.~\cite{mathew2024foundation} develop related foundation models for cardiovascular disease detection from digital-stethoscope biosignals, pre-trained self-supervised on large real-world datasets and adapted to multiple cardiac endpoints. These healthcare-specific PFMs illustrate how embedding domain structure and clinical objectives into pre-training can yield substantial gains over purely general-purpose time series models.

%% file: sections/5_lm4st.tex
\section{Large Models for Spatio-Temporal Data} \label{sec: Large Models for Spatio-Temporal Data}
In this section, we review recent advances in large models for spatio-temporal data, focusing on three major categories: spatio-temporal graphs, temporal knowledge graphs, and videos, all with broad real-world applications. We further organize the literature by model type and scope (see \shortautoref{fig:Taxonomy}).

\subsection{Spatio-Temporal Graphs} \label{sec: spatio-temporal graphs}
Individuals enter a planet imbued with inherent structure, where constituent elements engage in dynamic interactions across both spatial and temporal dimensions, culminating in a profound spatio-temporal composition. The conceptualization of this structural fabric within real-world problem-solving has been significantly enriched through the incorporation of \emph{spatio-temporal graphs} (STGs)~\cite{jin2023survey, jin2023spatio}. The application of STG forecasting, such as traffic prediction \cite{pan2020spatio,pan2021autostg}, air quality forecasting \cite{wang2020pm2,liang2022airformer}, stock price estimation \cite{qin2017dual,matsunaga2019exploring}, and human skeleton analysis \cite{shi2019skeleton,xu2023language}, has evolved into an indispensable tool for fostering informed decision-making and advancing sustainability objectives. In the era of deep learning, \emph{spatio-temporal graph neural networks} (STGNNs) \cite{jin2023survey,jin2023spatio} have emerged as the de-facto most popular approach for STG forecasting. They primarily use graph neural networks to capture spatial correlations among vertices and exploit other models (e.g., RNNs and CNNs) to learn temporal dependencies across different time steps.

Recently, the emergence of LLMs and PFMs has provided invaluable support to STGNNs within the domain of spatio-temporal representation learning. These models excel in processing and contextualizing textual data, equipping them with the capability to distill insights from diverse textual sources, ranging from news articles and social media content to reports. These insights can be seamlessly integrated into the spatio-temporal fabric, enhancing its contextual richness. Moreover, they facilitate the fusion of multiple modalities, including text, images, and structured data, thus amplifying the depth and breadth of spatio-temporal comprehension. With their capacity to generate human-interpretable explanations, these models foster transparency and reliability, especially in applications like urban planning or disaster response. Furthermore, they contribute to computational efficiency by encoding high-level information in concise representations, streamlining both training and inference processes.

\subsubsection{Large Language Models in Spatio-Temporal Graphs}
Recent work leverages LLMs for spatio-temporal forecasting by coupling them with graph-structured representations. STG-LLM~\cite{liu2024can} addresses the mismatch between text-trained LLMs and spatio-temporal graphs via an STG-Tokenizer that converts graph data into tokens and an STG-Adapter that interfaces these tokens with LLMs, yielding strong general-purpose forecasting. RePST~\cite{wang2025repst} reprograms pre-trained language models with a physics-aware decomposer and selective discrete reprogramming, enabling reasoning over disentangled spatial and temporal components. FSTLLM~\cite{jiangfstllm} targets few-shot forecasting by treating each time series channel as a graph node and using LLM-enhanced graph construction, combined with a downstream STGNN backbone, to improve robustness under data scarcity. ST-LINK~\cite{jeon2025st} further strengthens spatial and temporal modeling through Spatially-Enhanced Attention and a Memory Retrieval Feed-Forward Network. In the transportation domain, UrbanGPT~\cite{urbangpt}, GSF-LLM~\cite{wang2025gsf}, ST-LLM+~\cite{liu2025st}, and LLM-TFP~\cite{cheng2025llm} integrate graph encoders with partially frozen LLM backbones for traffic forecasting, jointly capturing spatial topology and temporal dynamics. Collectively, these models signal a shift from purely GNN-based designs toward LLM-centric, graph-aware architectures for spatio-temporal prediction. Their main advantage is the integration of language-level semantics with graph topology, while their practical boundary lies in the quality of graph construction, the stability of spatial dependencies, and the cost of adapting LLMs to large dynamic networks.

\subsubsection{Pre-Trained Foundation Models in Spatio-Temporal Graphs} \label{PFM_stg}
We are currently witnessing a proliferation of pre-trained machine learning models tailored for mastering STGs across an expansive spectrum of real-world applications. They significantly aid STG learning through knowledge transfer. In other words, they enable the transfer of learned representations, thereby improving the ability of STGs to capture intricate patterns, leading to more effective and efficient learning. In the subsequent parts, we introduce PFMs in both the general and specific domains.

\paragraph*{\textbf{General Purposes.}} Contrastive learning, a cornerstone of self-supervised representation learning, remains a powerful paradigm for advancing STG modeling. UMSST~\cite{zhong5297158unified} exemplifies this progress by introducing a two-stage pre-training strategy that captures both global and local spatio-temporal dependencies. It first learns holistic graph representations and then performs multi-subgraph contrastive pre-training to enhance representation transferability across downstream tasks such as forecasting and classification. Building upon these advances in pre-training, UniSTD~\cite{tang2025unistd} extends the scope from task-specific STG learning to a large-scale foundation model that unifies multiple spatio-temporal domains. Through a Transformer-based architecture with a rank-adaptive mixture-of-experts mechanism, UniSTD achieves robust generalization and cross-domain adaptability across a diverse range of forecasting and dynamic prediction tasks.

\paragraph*{\textbf{Climate.}} Recent work on domain-specific climate foundation models has shifted from task-specific predictors toward scalable, data-driven architectures. Pangu-Weather~\cite{bi2022pangu} introduces a 3D Earth-Specific Transformer trained on decades of reanalysis data, achieving fast, accurate medium-range global weather forecasts that rival and sometimes surpass traditional numerical weather prediction systems. ClimaX~\cite{nguyen2023climax} extends this idea to a general-purpose foundation model for weather and climate, using a Transformer pre-trained on heterogeneous CMIP6 datasets so it can be efficiently fine-tuned across variables, spatial resolutions, and time scales. Building on earlier Fourier-operator-based models, FourCastNet-3~\cite{bonev2025fourcastnet} adopts a geometric, sphere-aware architecture with probabilistic ensembles, delivering calibrated long-range forecasts and combining strong skill with substantial gains in computational efficiency. Together, these models exemplify a transition toward versatile climate PFMs capable of both precision and scalability.

\paragraph*{\textbf{Transportation.}} In the domain of traffic flow forecasting, TrafficBERT \cite{jin2021trafficbert} leverages key characteristics inspired by BERT~\cite{devlin2018bert}. It employs a bidirectional transformer structure akin to BERT, allowing it to predict overall traffic flow instead of individual time steps. In contrast to conventional models that require distinct training for each specific road, TrafficBERT enhances model generalization by pre-training it using data from various roads. Furthermore, Wang et al. introduce the Transportation Foundation Model (TFM) \cite{wang2023building}, which incorporates traffic simulation into the realm of traffic prediction. TFM leverages graph structures and dynamic graph generation algorithms to adeptly capture the intricate dynamics and interactions among participants within transportation systems. This data-driven and model-free simulation approach effectively tackles the long-standing challenges posed by conventional systems in terms of structural complexity and model accuracy, laying a robust foundation for addressing intricate transportation issues with real-world data.

\subsection{Temporal Knowledge Graphs} \label{sec: temporal knowledge graphs}
A host of applications such as search engines, question-answering systems, conversational dialogue systems, and social networks require reasoning over underlying structured knowledge. In particular, \emph{knowledge graphs} (KGs)~\cite{suchanek07yogo,vrande2014wikidata} have gained much attention as an important model for studying complex multi-relational settings over such knowledge. KGs represent facts (events), usually extracted from text data, in the form of triples $(s,p,o)$ where $s$ and $o$ denote subject and object entities respectively, and $p$ as a relation type means predicate. However, in the real world, knowledge evolves continuously, inspiring the construction and application of the \emph{temporal knowledge graphs} (TKGs)~\cite{trivedi2017knowevolve,trivedi2019dyrep}, where the fact extends from a triple $(s, p, o)$ to a quadruple with a timestamp $t$, i.e., $(s, p, o, t)$. By effectively capturing temporal dependencies across facts in addition to the structural dependencies, TKGs help improve the understanding on the behavior of entities and how they contribute to the generation of facts over time.

Recently, LLMs~\cite{OpenAI2023GPT4TR} have demonstrated strong performance on various text-reasoning tasks, motivating their use in TKG reasoning. Based on the tasks performed, existing LLM-related TKG models can be broadly divided into two categories: \emph{forecasting} and \emph{completion}. For forecasting, Xia \textit{et al.}~\cite{xia2024chain} propose Chain-of-History (CoH) reasoning, which incrementally incorporates higher-order historical facts into LLM prompts to provide rich yet compact temporal context for event prediction. Furthermore, LLM-DA~\cite{wang2024large} takes a complementary route by using LLMs to extract and dynamically adapt temporal logic rules as new events arrive, enabling interpretable and up-to-date temporal reasoning without repeated fine-tuning. LATE~\cite{zhang2024harnessing} further enhances robustness by combining structural, temporal, and textual signals through multi-view prompting, improving factual consistency across diverse temporal settings. Building on these advances, G2S~\cite{bai2025g2s} adopts a general-to-specific strategy that first trains LLMs on anonymized temporal structures to capture domain-agnostic patterns, then injects scenario-specific entity and relation information, while AnRe~\cite{tang2025anre} designs a training-free analogical replay mechanism that retrieves and replays similar historical event chains so LLMs can exploit both semantic understanding and structural evolution for forecasting.

For completion, LLM-based and embedding-based approaches have also begun to converge. Luo \textit{et al.}~\cite{luo2024chain} extend the Chain-of-History idea to temporal knowledge graph completion by formulating temporal link prediction as event generation along historical chains and applying parameter-efficient fine-tuning. This shows that LLMs can jointly handle interpolation and extrapolation of temporal facts. TeRDy~\cite{liu2025terdy} addresses temporal relation dynamics via frequency decomposition, modeling long-term trends and short-term fluctuations of relations with low- and high-frequency components and injecting them into relation–timestamp embeddings, thereby providing a strong, general-purpose TKGC backbone. More recently, \cite{ong2025dynamic} integrates LLMs with graph structural representations to address the challenge of emerging entities and relations in evolving TKGs. By jointly leveraging textual semantics and temporal graph context, it enhances the model’s ability to infer missing links in dynamic environments. Overall, LLM-based TKG methods progress from prompt-based historical reasoning to retrieval, rule adaptation, and structure-aware completion, but their effectiveness remains constrained by incomplete facts, sparse timestamps, and the difficulty of aligning symbolic graph dynamics with continuous temporal changes.

\subsection{Videos} \label{sec: video data}
Video refers to the digital representation of visual information, typically composed of a sequence of images or frames that collectively convey motion and temporal changes. These data have become ubiquitous in various real-world applications, including surveillance systems, entertainment platforms, social media, and driver assistance systems~\cite{vishwakarma2013survey}. Conventional deep learning approaches for video understanding primarily involve two key paradigms: (1) 2D CNNs in which each video frame is processed separately by 2D convolutions and then aggregated along the time axis at the top of the network~\cite{karpathy2014large,wang2018temporal}; (2) 3D CNNs which learn spatio-temporal video representation via 3D convolutions by aggregating spatial and temporal features~\cite{du2015c3d,xu2017r}. Recently, Transformers have also been widely used in modeling spatio-temporal dependencies for video recognition~\cite{fan2021multiscale,arnab2021vivit} by virtue of their capability in capturing long-range dependencies.

However, recent advances in LLMs and PFMs have paved the way for enhanced video understanding by leveraging the inherent multimodal nature of videos. These models, such as OpenAI's CLIP~\cite{clip} and DALL-E~\cite{ramesh2022hierarchical}, can effectively extract rich contextual information from video data by jointly processing visual and textual modalities, enabling a more comprehensive understanding of complex scenes and events. Furthermore, these models have the potential to facilitate efficient transfer learning across diverse domains, leading to improved generalization and robustness in video analysis tasks. We will discuss LLMs and PFMs for video understanding in the following parts.

\subsubsection{Large Language Models for Video Data}
The emergence of large multimodal models marks a shift in video understanding research — from narrowly defined recognition tasks toward comprehensive, reasoning-driven interpretation of dynamic visual data. Recent advancements in LLMs have facilitated the application of sequence reasoning capabilities, derived from pre-trained LLMs in natural language processing, to various video processing tasks as demonstrated in LAVENDER \cite{li2023lavender}. Furthermore, Video-ChatGPT \cite{maaz2023video} is a multimodal model tailored for video-based conversation. It combines a visual encoder adapted for processing video content with LLMs, enabling the model to comprehend and generate in-depth discussions related to videos. In a parallel line of research, Zhang et al. propose Video-LLaMA~\cite{zhang2023video}, another multi-modal framework that operates independently of textual data. Video-LLaMA empowers Llama~\cite{touvron2023llama} with visual and audio information via Q-Former to endow it with the ability to comprehend both visual and auditory information present in videos. To ensure alignment between the output of the visual and audio encoders and LLM's embedding space, Video-LLaMA is trained on an extensive corpus of video/image-caption pairs, as well as visual-instruction-tuning datasets of moderate size but with superior quality. Video-LLaMA-3~\cite{zhang2025videollama} further advances this framework by enhancing multimodal alignment and temporal reasoning, enabling more coherent understanding of long-form and dynamic video content. 

Around the same period, other approaches began exploring complementary directions in long-context reasoning, causal inference, and memory-efficient modeling for extended video understanding. CaKE-LM~\cite{su2023language} demonstrates that language models can act as causal knowledge extractors, enabling zero-shot video question answering by leveraging intrinsic world knowledge within LLMs. MovieChat~\cite{song2023moviechat} introduces a sparse memory mechanism that replaces dense token processing with selective memory representations, effectively scaling LLMs to handle videos exceeding 10,000 frames. Expanding these ideas, MA‑LMM~\cite{he2024ma} introduces a memory-augmented mechanism for long-term video understanding, enabling the model to retain contextual information over extended temporal spans, while Apollo~\cite{zohar2025apollo} explores video understanding within large multimodal models through comprehensive scaling studies and unified pretraining strategies, improving generalization across diverse video tasks. In parallel, video‑SALMONN‑o1~\cite{sun2025video} incorporates reasoning-enhanced audio-visual modeling by combining structured cognitive reasoning with multimodal fusion to improve video question answering and inference. Moving beyond understanding, ProgGen~\cite{tang2025programmatic} applies programmatic reasoning to enable LLMs to perform video prediction and dynamic event modeling, demonstrating the potential of integrating generative reasoning into temporal visual tasks. Moreover, Valley~\cite{luo2023valley} employs a spatio-temporal pooling strategy to extract a unified visual encoding from video and image inputs. It also curates a vast corpus of vision-text pairs for pre-training purposes, and subsequently generates a multi-task instruction-following video dataset, aided by ChatGPT for conversation design. 

Building on the long-form and generative push, two very recent works push the frontier of video-LLMs further. Video-XL~\cite{shu2025video} addresses the challenge of hour-scale video understanding by introducing a visual summarization token mechanism that compresses long visual sequences into compact representations for efficient processing. Simultaneously, BOLT~\cite{liu2025bolt} proposes a training-free frame selection strategy that improves long-form video VLM performance by choosing query-relevant frames during inference, thus boosting performance without additional model training. Together, these works illustrate the current frontier: handling drastically longer video inputs, compressing and selecting relevant information, and maintaining reasoning capability across extended temporal horizons.

Beyond general-purpose video understanding, researchers have begun to tailor large vision–language models for specific real-world domains. In the traffic domain, Qasemi et al. introduce an innovative automatic captioning method called TRIVIA~\cite{qasemi2023traffic}, which serves as a form of weak supervision to incorporate traffic-domain knowledge into extensive video-language models. In the sports domain, Rao et al. present MatchVision~\cite{rao2025towards}, a large-scale foundation model trained on the SoccerReplay-1988 dataset that supports unified soccer video understanding across event recognition, commentary generation, and foul detection. Jiang et al.~\cite{jiang2025domain} further refine this approach by fine-tuning a general video–language model through curriculum-style adaptation on thousands of soccer clips, which leads to substantial gains in domain-specific action recognition and video question answering accuracy. These studies demonstrate that adapting multi-modal LLMs to structured, domain-centric contexts produces more accurate and semantically grounded video understanding.

\subsubsection{Pre-Trained Foundation Models for Video Data}
In recent years, language, image-based vision, and multi-modal pre-training methodologies have significantly converged. Following this trend, recent research has pioneered various video-oriented pre-training strategies in a contrastive or generative way \cite{ye2022hitea,he2023vlab,zhong2023stoa}, leading to a series of video or video-language foundation models for video processing \cite{wang2022omnivl,wang2023paxion,xu2023mplug,xu2023youku}. Specifically, Wang et al. first present a vision-language foundation model called OmniVL \cite{wang2022omnivl}, which unifies image-language and video-language modeling. It can naturally support a wide range of tasks, including visual-only tasks, cross-modal alignment tasks, and multi-modal tasks. Furthermore, mPLUG-2 \cite{xu2023mplug} is a new Transformer framework, which allows for the utilization of various combinations of modules for both uni-modal and cross-modal tasks. To achieve this, mPLUG-2 shares universal modules while separating modality-specific ones, effectively addressing the issue of modality entanglement and offering flexibility in selecting various modules for diverse understanding and generation tasks across all modalities, encompassing text, image, and video. Building on the concept of mPLUG-2, Xu et al. further devise a modularized decoder-only model named mPLUG-video \cite{xu2023youku} with a restricted number of trainable parameters. This model is built upon a frozen pre-trained LLM and Youku-mPLUG\footnote{The first public Chinese video-language pre-training dataset \protect\cite{xu2023youku}.}. Building on these developments, mPLUG-Owl2 ~\cite{ye2024mplug} introduces a modality collaboration mechanism that strengthens interactions among text, image, and video representations within a unified multimodal framework, improving cross-modal alignment and reasoning. InternVideo~\cite{wang2024internvideo2} further scales video foundation models by combining masked video modeling, cross-modal contrastive learning, and next-token prediction, advancing long-form video comprehension and multimodal generalization. MSTA~\cite{chen2025efficient} complements these large-scale models with an efficient transfer learning approach that uses lightweight adapter modules and spatio-temporal consistency constraints to enable effective adaptation with minimal fine-tuning.

\subsection{Methodological Convergence and Cross-Domain Insights}

Although time series and spatio-temporal data differ in structure and modality, LM4TS and LM4STD share the goal of adapting large models to preserve temporal order, contextual information, and domain semantics. Three challenges are common. First, temporal observations must be tokenized into model-readable units: LM4TS maps sequences into patches, symbols, summaries, or prompt-aligned embeddings \cite{xue2022promptcast,jin2023timellm,ansari2024chronos,goswami2024moment}, whereas LM4STD further augments tokens with topology, entity relations, trajectories, or visual frames for STGs, TKGs, and videos \cite{liu2024can,liu2025st,xia2024chain,wang2024large,shu2025video,liu2025bolt}. Second, both areas must handle long-range dependencies under limited context budgets. LM4TS often uses patching, decomposition, or lightweight adaptation \cite{nie2022time,woo2024unified}, while LM4STD relies on graph summarization, spatio-temporal pooling, memory modules, or frame selection to manage expansion over nodes, entities, regions, or frames \cite{liu2025st,shu2025video,liu2025bolt}. Third, both require external knowledge and modality alignment: LM4TS mainly uses prompts, calendar features, domain descriptions, or historical retrieval \cite{xue2022promptcast,jin2023timellm}, whereas LM4STD can additionally exploit road networks, entity relations, trajectories, and video-language alignment \cite{liu2024can,wang2024internvideo2}. These subfields also have complementary trade-offs: LM4TS is simple and scalable but may lose fine-grained numerical information; STG methods exploit topology but depend on graph quality; TKG methods support interpretable relational reasoning but suffer from incompleteness and sparsity; and video models capture rich visual-temporal context but incur high token and memory costs. Thus, graph tokenization, spatio-temporal pooling, retrieval, video memory, and frame selection can inspire structured representations and compression for time series analysis, while prompt-based and adapter-based transfer from LM4TS can reduce the cost of adapting large models to STGs, TKGs, and videos.

%% file: sections/6_resource_and_application.tex
\section{Resources and Applications} \label{sec:resources}

In this section, we summarize the common datasets, models, and tools in various applications related to time series and spatio-temporal data, which are listed in \shortautoref{tab:model_and_data_simplified}.

\input{tables/resources_simplified}

\subsection{Traffic Application}
Traffic flow forecasting has become a pivotal concern in the development of Intelligent Transportation Systems (ITS). The utilization of time series and spatio-temporal data aids in creating more accurate and adaptive predictive models. Several datasets have emerged as benchmarks in the field. For instance, \textit{METR-LA} \cite{li2018dcrnn_traffic} is a notable dataset that encompasses traffic speed data procured from loop detectors stationed across the LA County road network. Similarly, \textit{PEMS-BAY} \cite{li2018dcrnn_traffic} and \textit{PEMS04} \cite{doi:10.3141/1748-12} are datasets derived from the California Transportation Agencies' Performance Measurement System, offering exhaustive insights into the traffic speed and flow in the Bay Area over distinct periods. Another significant dataset is \textit{SUTD-TrafficQA} \cite{Xu_2021_CVPR}, which comprises 10,080 real-world videos annotated with 62,535 question-answer pairs, capturing both stationary and ego-view perspectives. Furthermore, \textit{LargeST} \cite{liu2023largest} is the first traffic dataset for large-scale traffic forecasting, encompassing over 8,000 sensors over road networks with a period of five years. For effective analysis and simulation, a number of tools are at the disposal of researchers. For instance, \href{https://sumo.dlr.de/docs/Tutorials/index.html}{\textit{SUMO}} stands out as an open-source traffic simulator, and \href{https://www.safegraph.com/blog/safegraph-partners-with-dewey}{\textit{SafeGraph Data for Academics}} offers a platform to host anonymized real-world human mobility data.

\subsection{Healthcare Application}
Time series forecasting in healthcare has burgeoned due to its practical implications, despite inherent methodological challenges. Within medical contexts, forecasting models have been pivotal in predicting disease progression, estimating mortality rates, and evaluating time-dependent risks. To date, several datasets stand out. For instance, \textit{PTB-XL}~\cite{ptb-xl} comprises 21,837 clinical 12-lead ECG of 10-second duration from 18,885 patients, categorized into diagnosis, form, and rhythm, with 71 different statements in total. \textit{NYU Datasets}~\cite{jiang2023health} include NYU Notes, NYU Notes–Manhattan, NYU Notes–Brooklyn, spanning 10 years of unlabelled inpatient clinical notes. Fine-tuning datasets like "NYU Readmission", "NYU Mortality", among others, contain specific labels. \textit{UF Health clinical corpus}~\cite{yang2022large} is an aggregation of clinical narratives from UF Health IDR, MIMIC-III corpus, PubMed collection, and Wikipedia articles, resulting in a more than 90 billion-word corpus. \textit{MIMIC-III}~\cite{mimic} is a public dataset with ICD-9 codes, vital signs, medications, and other critical patient data from intensive care units. In addition, a variety of model checkpoints and toolkits have emerged, tailored to healthcare applications, such as \textit{NYUTron}~\cite{jiang2023health}, \textit{BioBERT}~\cite{lee2020biobert}, \textit{BlueBERT}~\cite{peng2019transfer}, and \textit{ClinicalBERT}~\cite{huang2019clinicalbert}.

\subsection{Weather Application}
Weather modeling primarily involves predicting atmospheric conditions, which are fundamental to a vast array of daily decisions and economic considerations. The analysis of time series in weather forecasting has greatly benefited from significant advancements in data collection and computational modeling. For instance, datasets such as \textit{AvePRE}, \textit{SurTEMP}, and \textit{SurUPS} provided by NASA~\cite{chen2023prompt} shed light on the hourly changes of 12 weather parameters, captured by an extensive network of ground-based instruments. Moreover, \textit{CMIP6}~\cite{eyring2016overview} is an international consortium focusing on the evaluation of diverse global climate models. In the realm of recent foundational models, \textit{Pangu-Weather}~\cite{bi2022pangu} is renowned for its rapid and precise global forecasts, while \textit{GraphCast}~\cite{lam2022graphcast}, which incorporates GNNs, is noted for its adaptability in various prediction tasks.

\subsection{Finance Application}
Time series analysis in finance presents significant challenges, as it requires modeling both linear and non-linear historical interactions for future prediction. Common applications include predicting buy/sell signals and stock price shifts. \textit{EDT}~\cite{zhou-etal-2021-trade} and \textit{NASDAQ-100}~\cite{qin2017dual} are two popular datasets: the former supports corporate event detection and text-based stock prediction, while the latter contains NASDAQ-100 stock prices retrieved from Yahoo Finance. \textit{Finance (Employment)}~\cite{aa-forecast} captures employment changes based on one million active US employees during the COVID-19 pandemic and incorporates state-level policies as extreme-event markers. \textit{StockNet}~\cite{xu2018stock} covers 88 stocks over two years, combining tweets and historical stock prices. Recently, financial LLMs such as \textit{WeaverBird}~\cite{xue2023weaverbird} and \textit{FinGPT}~\cite{yang2023fingpt} have emerged for financial prediction and decision-making.

\subsection{Video Application}
Video question answering (VQA) aims to answer natural language questions using video content, with models expected to generate answers that accurately reflect the depicted scenes. The domain also includes video quality assessment and future video content prediction. Datasets such as \textit{TGIF-QA}~\cite{jang-IJCV-2019}, which provides 165K GIF-based QA pairs, and \textit{WebVid}~\cite{bain2021frozen}, which contains 10 million annotated video clips, are notable examples. \textit{MSR-VTT (Microsoft Research Video to Text)}~\cite{xu2016msr} supports video captioning with 10,000 clips across 20 categories, while \textit{COCO (Microsoft Common Objects in Context)}~\cite{lin2014microsoft} supports object detection, segmentation, and captioning with 328K images. Contrastive language-image pre-training \href{https://github.com/openai/CLIP}{\textit{(CLIP)}} uses natural language supervision to train image representations. \textit{BLIP}~\cite{li2022blip} introduces a bootstrapping method for noisy web data in vision-language pre-training, and \textit{ViLBERT (Vision-and-Language BERT)}~\cite{lu2019vilbert} extends BERT to jointly process visual and textual inputs.

\subsection{Event Prediction Application}
Event sequences, also known as asynchronous time series with irregular timestamps, play a crucial role in various real-world domains, including finance, online shopping, and social networks. \textit{Amazon}~\cite{xue2023easytpp} and \textit{Taobao}~\cite{xue2023easytpp} are two representative datasets of time-stamped user product-review behavior on e-commerce platforms. \textit{Retweet}~\cite{xue2023easytpp}, \textit{StackOverflow}~\cite{xue2023easytpp} and  \textit{Taxi}~\cite{xue2023easytpp} are three other commonly used event sequence datasets, which contain user retweet, user question-answer, and user taxi pickup event sequences, respectively. In terms of useful tools, \textit{Tick}~\cite{bacry2017tick} is a well-established library focused on statistical learning for classical TPPs, while \textit{EasyTPP}~\cite{xue2023easytpp} is the first open-source central repository offering a comprehensive suite of research resources, including data, models, evaluation programs, and documentation, in the field of neural event sequence modeling.

\subsection{Other Applications}
Beyond the applications above, time series forecasting, classification, and anomaly detection are widely used in electricity, cloud computing, and other sectors. Several benchmark datasets support these tasks, including \textit{ETT (Electricity Transformer Temperature)}~\cite{haoyietal-informer-2021} for transformer temperature forecasting, \textit{Alibaba Cluster Trace}~\cite{cheng2018characterizing} for cloud workload modeling, \textit{TSSB}~\cite{clasp2021} for time-series segmentation, and the \textit{UCR Time Series Classification Archive}~\cite{dau2019ucr} for classification. These resources cover diverse temporal patterns, sampling frequencies, and application settings, making them useful for evaluating the generalization of temporal models. For reproducible research, tools such as \href{https://github.com/chengtan9907/OpenSTL}{\textit{OpenSTL}}~\cite{tan2023openstl},  \href{https://github.com/salesforce/Merlion}{\textit{Merlion}}~\cite{bhatnagar2021merlion}, \href{https://unit8co.github.io/darts/}{\textit{darts}}~\cite{JMLR:v23:21-1177}, and \href{https://github.com/benedekrozemberczki/pytorch_geometric_temporal}{\textit{PyTorch Geometric Temporal}}~\cite{rozemberczki2021pytorch} provide benchmarks for time series and spatio-temporal modeling. \\

%% file: tables/resources_simplified.tex
\begin{table*}
    \caption{A summary of common dataset resources in different applications.}
    \label{tab:model_and_data_simplified}
    \centering
    \renewcommand\arraystretch{0.9}
    \resizebox{\textwidth}{!}{
    \begin{tabular}{c|ccccc} 
    \toprule\toprule	
	Applications & Dataset & Scale / Data Type & Timeframe & Source & Citations \\ 
    \midrule
	
	\multirow{5}{*}{Traffic} 
 	
        & \href{https://zenodo.org/record/5146275}{METR-LA} 
        & 34,272 time steps; traffic sensor series 
        & Mar to Jun 2012 
        & \cite{li2018dcrnn_traffic} 
        & \cite{shao2022pre,jin2022multivariate}
        \\
        
        & \href{https://zenodo.org/record/5146275}{PEMS-BAY} 
        & 52,116 time steps; traffic sensor series 
        & Jan 1 to May 31, 2017 
        & \cite{li2018dcrnn_traffic} 
        & \cite{shao2022pre,jin2022multivariate} 
        \\ 
        
        & \href{https://github.com/Davidham3/ASTGCN}{PEMS04} 
        & 16,992 time steps; traffic sensor series 
        & Jan 1 to Feb 28, 2018 
        & \cite{doi:10.3141/1748-12} 
        & \cite{shao2022pre}
        \\
        
        & \href{https://github.com/sutdcv/SUTD-TrafficQA}{SUTD-TrafficQA}
        & 10,080 videos; 62,535 questions 
        & - 
        & \cite{Xu_2021_CVPR} 
        & \cite{qasemi2023traffic}
        \\ 
        
        & \href{https://github.com/liuxu77/LargeST}{LargeST}
        & 525,888 frames; 8,600 sensors 
        & 2017 to 2021 
        & \cite{liu2023largest} 
        & \cite{liu2023largest}
        \\
    
	\midrule
	
	\multirow{4}{*}{Healthcare} 
	
        & \href{https://physionet.org/content/ptb-xl/1.0.3/}{PTB-XL} 
        & 21,837 records; ECG signals 
        & Oct 1989 to Jun 1996 
        & \cite{ptb-xl} 
        & \cite{mehari2022self,wang2021automated,hu2022transformer} 
        \\

        & \href{https://datacatalog.med.nyu.edu/dataset/10633}{NYUTron} 
        & 387,144 records; clinical text/EHR 
        & Jan 2011 to May 2020 
        & \cite{jiang2023health} 
        & \cite{qi2023evaluating,han2023medical}
        \\
        
        & \href{https://catalog.ngc.nvidia.com/orgs/nvidia/teams/clara/models/gatortron_og}{UF Health clinical corpus} 
        & 2M records; clinical text 
        & 2011 to 2021 
        & \cite{yang2022large} 
        & \cite{thirunavukarasu2023large,moor2023foundation} 
        \\
 
        & \href{https://physionet.org/content/mimiciii/1.4/}{MIMIC-III} 
        & 112,000 records; ICU/EHR data 
        & 2001 to 2012 
        & \cite{mimic} 
        & \cite{peng2019transfer,zhang2022shifting,xu2021federated}
        \\

        \midrule 
	
	\multirow{4}{*}{\makecell[c]{Weather}} 
        
        & \href{https://disc.gsfc.nasa.gov/}{AvePRE} 
        & 88 satellite series; meteorological data 
        & Apr 1, 2012 to Feb 28, 2016 
        & \cite{chen2023prompt} 
        & \cite{chen2023mask} 
        \\

        & \href{https://disc.gsfc.nasa.gov/}{SurTEMP} 
        & 525 satellite series; surface temperature data 
        & Jan 3, 2019 to May 2, 2022 
        & \cite{chen2023prompt} 
        & \cite{chen2023mask} 
        \\

        & \href{https://disc.gsfc.nasa.gov/}{SurUPS} 
        & 238 satellite series; meteorological data 
        & Jan 2, 2019 to Jul 29, 2022 
        & \cite{chen2023prompt} 
        & \cite{chen2023mask} 
        \\

        & \href{https://cds.climate.copernicus.eu/cdsapp\#!/dataset/projections-cmip6?tab=overview}{CMIP6} 
        & Global gridded climate data; 1 vertical level 
        & 1850--2014 for historical and 2015--2100 for SSP expt. 
        & \cite{eyring2016overview} 
        & \cite{lee2021future,hassani2021global,jones2022global} 
        \\

        \midrule

        \multirow{4}{*}{\makecell[c]{Finance}} 
        
        & \href{https://github.com/ashfarhangi/AA-Forecast/tree/main/dataset}{Finance (Employment)} 
        & 15,321 records; employment/finance data 
        & during COVID-19 
        & \cite{aa-forecast} 
        & \cite{wu2023symphony} 
        \\
        
        & \href{https://github.com/yumoxu/stocknet-dataset}{StockNet} 
        & 88 stocks; price/news data 
        & 2014/01/01 to 2016/01/01 
        & \cite{xu2018stock} 
        & \cite{sawhney2020deep,xie2023pixiu,jiang2021applications,zou2022astock,xie2022word,feng2018enhancing} 
        \\
        
        & \href{https://github.com/Zhihan1996/TradeTheEvent/tree/main/data\#edt-dataset}{EDT} 
        & 303,893 news articles; minute-level timestamps 
        & 2020/03/01 to 2021/05/06 
        & \cite{zhou-etal-2021-trade} 
        & \cite{liu2023fingpt,reneau2023feature} 
        \\
        
        & \href{https://cseweb.ucsd.edu/~yaq007/NASDAQ100_stock_data.html}{NASDAQ-100} 
        & 104 corporations; one-minute stock series 
        & July 26, 2016 to April 28, 2017 
        & \cite{qin2017dual} 
        & \cite{liu2020finrl} 
        \\

        \midrule

        \multirow{4}{*}{\makecell[c]{Video}} 
        
        & \href{https://github.com/YunseokJANG/tgif-qa}{TGIF-QA} 
        & 165K QA pairs; GIF/video data 
        & - 
        & \cite{jang-IJCV-2019} 
        & \cite{xu2023multi,maaz2023video,li2023videochat,xu2023mplug} 
        \\
        
        & \href{https://drive.google.com/file/d/1pWym3bMNW_WrOZCi5Ls-wKFpaPMbLOio/view}{MSR-VTT} 
        & 10,000 video clips; video-text data 
        & 41.2 hours 
        & \cite{xu2016msr} 
        & \cite{he2023vlab,chen2023vast,zeng2022socratic} 
        \\
        
        & \href{https://maxbain.com/webvid-dataset/}{WebVid} 
        & 10M video clips; web video-text data 
        & 13K hours 
        & \cite{bain2021frozen} 
        & \cite{wang2023videofactory,luo2023videofusion} 
        \\
        
        & \href{https://cocodataset.org/\#home}{COCO} 
        & 328K images; image-text data 
        & - 
        & \cite{lin2014microsoft} 
        & \cite{li2022image,wang2023internimage} 
        \\

        \midrule

        \multirow{5}{*}{\makecell[c]{Event Prediction}} 
       
        & \href{https://github.com/ant-research/EasyTemporalPointProcess}{Amazon} 
        & 5.2K event sequences; temporal point process
        & January, 2008 to October, 2018 
        & \cite{xue2023easytpp} 
        & \cite{xue2022hypro,shi2023language,xue2023prompttpp} 
        \\
        
        & \href{https://github.com/ant-research/EasyTemporalPointProcess}{Taobao} 
        & 4.8K event sequences; temporal point process 
        & November 25, 2017 to December 03, 2017 
        & \cite{xue2023easytpp} 
        & \cite{xue2022hypro,xue2023prompttpp} 
        \\
        
        & \href{https://github.com/ant-research/EasyTemporalPointProcess}{Retweet} 
        & 5.2K event sequences; temporal point process
        & - 
        & \cite{xue2023easytpp} 
        & \cite{mei2017neural,zhang2020self,zuo2020transformer,wang2023ep} 
        \\
        
        & \href{https://github.com/ant-research/EasyTemporalPointProcess}{StackOverflow} 
        & 2.2K event sequences; temporal point process 
        & 2 years 
        & \cite{xue2023easytpp} 
        & \cite{xue2021graphpp,xue2022hypro,xue2023prompttpp} 
        \\
        
        & \href{https://github.com/ant-research/EasyTemporalPointProcess}{Taxi} 
        & 2K event sequences; temporal point process
        & 1 year 
        & \cite{xue2023easytpp} 
        & \cite{mei2020neural,xue2022hypro} 
        \\

        \midrule
        
        \multirow{6}{*}{\makecell[c]{Others}} 
       
        & \href{https://github.com/zhouhaoyi/ETDataset}{ETT} 
        & 7-feature time series; hourly/15-minute granularity 
        & 2 years 
        & \cite{haoyietal-informer-2021} 
        & \cite{Zeng2022AreTE,liu2022SCINet} 
        \\
        
        & \href{https://github.com/Mcompetitions/M4-methods}{M4} 
        & 100,000 series; multi-frequency time series 
        & - 
        & \cite{MAKRIDAKIS2018802} 
        & \cite{NBeatsPRemy,agtabular,bhatnagar2021merlion} 
        \\
       
        & \href{https://archive.ics.uci.edu/dataset/235/individual+household+electric+power+consumption}{Electricity} 
        & 2,075,259 observations; household power series 
        & 4 years 
        & \cite{misc_individual_household_electric_power_consumption_235} 
        & \cite{liu2022SCINet,DU2023SAITS,lai2023lightcts} 
        \\
        
        & \href{https://github.com/alibaba/clusterdata}{Alibaba Cluster Trace} 
        & 1,313 machines; cluster trace data 
        & 12 hours 
        & \cite{cheng2018characterizing} 
        & \cite{mommessin2022affinityaware} 
        \\
        
        & \href{https://github.com/ermshaua/time-series-segmentation-benchmark}{TSSB} 
        & 75 series; segmentation benchmark 
        & - 
        & \cite{clasp2021} 
        & \cite{deldari2020espresso,wss2022} 
        \\
        
        & \href{https://www.cs.ucr.edu/~eamonn/time_series_data_2018/}{UCR TS Classification Archive} 
        & 128 datasets; time-series classification archive 
        & 2002 to 2018 
        & \cite{dau2019ucr}
        & \cite{10.1007/978-3-031-09037-0_53,cheng2022classification,schaefer2023weasel2} 
        \\
        
    \bottomrule\bottomrule
    \end{tabular}
    }
\end{table*}

%% file: sections/7_future_direction.tex
\vspace{-4.5mm}

\section{Outlook and Future Opportunities} \label{sec:future directions}
In this section, we discuss the potential limitations of current research and highlight the top six future research directions aimed at developing more powerful, transparent, and reliable large models for temporal data.

\subsection{Theoretical Analysis of Large Models}
Large models, especially LLMs, were primarily developed for natural language processing, but their ability to model long-range sequential dependencies has motivated their use in time series and spatio-temporal tasks. However, the theoretical basis for this transfer remains limited. It is still unclear when linguistic representations, tokenization schemes, or attention mechanisms can effectively capture temporal patterns, and when they may fail due to distribution shift, numerical precision loss, or modality mismatch. Future work should develop deeper theoretical analyses of large models for temporal data, including the similarities and differences between language and temporal sequences, the conditions under which pre-trained representations can transfer, and the task-specific limits of large models in forecasting, anomaly detection, classification, and spatio-temporal reasoning.

\subsection{Development of Multimodal Models}
Real-world time series and spatio-temporal data are accompanied by supplementary information, such as text descriptions, news articles, or tweets. This is especially useful in economics and finance, where forecasting can combine textual signals with numerical time series. LLMs can therefore be adapted to learn joint representations that capture both temporal dynamics and modality-specific information. Since different modalities may have different temporal resolutions, future models should better align multi-resolution temporal signals to leverage cross-modal information. Recent work on networking systems for video anomaly detection further suggests that cross-domain temporal modeling should consider deployability constraints, including edge-cloud collaboration, latency, privacy, and robustness, in addition to algorithmic accuracy~\cite{liu2025networking}.

\subsection{Continuous Learning and Adaptation}
Real-world applications often involve non-stationary temporal patterns, making it important for large models to adapt continuously without catastrophic forgetting. While online learning, concept drift adaptation, and evolving-pattern modeling have been studied in conventional machine learning, their integration with large temporal models remains under-explored. Cross-domain transfer is also important when source-domain data cannot be shared due to privacy or deployment constraints; recent source-free foundation-model-enabled state-of-health estimation for lithium-ion batteries shows that intelligent adapter mapping can transfer temporal representations across battery domains, suggesting a promising direction for privacy-preserving and domain-adaptive PFMs in industrial applications~\cite{qin2026source}.

\subsection{Interpretability and Explainability}
Understanding why a model produces a particular forecast or prediction is crucial for time series analysis, especially in high-stakes domains such as healthcare and finance. However, the internal mechanisms of LLMs remain difficult to interpret, making it important to develop theoretical frameworks and practical tools for analyzing what they learn from temporal data and how they generate predictions. Improved interpretability can provide rationales for forecasts, anomaly detection, and classification decisions. A promising direction is to integrate explicit temporal reasoning into large models, enabling counterfactual analysis, causal inference, root cause analysis, scenario simulation, and intervention planning, where understanding not only \emph{what} will happen but also \emph{why} it happens is essential.

\subsection{Privacy and Adversarial Attacks on Large Models}
Temporal data can be highly sensitive, especially in applications like healthcare and finance. When LLMs are trained or fine-tuned on such data, they can memorize specific details, posing the risk of leaking private information. Privacy-preserving techniques, such as differential privacy and federated learning, offer promising directions for ensuring data privacy while still benefiting from the power of LLMs for time series and spatio-temporal analysis.

\subsection{LLM Agents and Decision Making Process in Time Series Systems}
LLMs can act as agents that continuously observe temporal or spatio-temporal signals and select actions. In such systems, prediction errors can accumulate through feedback into the environment, making reliable decision policies critical. Real-world temporal data often exhibit non-stationarity, regime shifts, and rare but high-impact events. Under these distribution shifts, LLM-based agents may issue overconfident or unsafe decisions, especially when policies are learned from noisy, biased, or adversarially manipulated trajectories. This motivates research on robust, uncertainty-aware decision-making for LLM agents in time series and spatio-temporal settings, including out-of-distribution detection and abstention, conservative or risk-sensitive policies, safe exploration, and human- or rule-in-the-loop oversight to prevent harmful cascades and ensure alignment with safety and performance requirements.

%% file: sections/8_conclusion.tex
\section{Conclusion} \label{sec:conclusion}
We present an extensive and up-to-date survey of large models for the analysis of time series and spatio-temporal data. Our aim is to offer a fresh perspective on this dynamic field by introducing a novel taxonomy that categorizes the reviewed models. We summarize the most prominent techniques in each category, examine their strengths and limitations, and illuminate promising avenues for future research. Within this exciting topic, the scope for groundbreaking investigations is boundless. This survey will serve as a catalyst for stimulating further curiosity and fostering an enduring passion for research in the realm of large models for time series and spatio-temporal data analysis.